%% file: main.tex
\documentclass[11pt]{article} 
\usepackage{smile}
\usepackage{epstopdf}
\usepackage{wrapfig}
\usepackage[colorlinks, linkcolor=blue, anchorcolor=blue, citecolor=blue]{hyperref}
\usepackage[margin=1in]{geometry}
\usepackage[normalem]{ulem}
\usepackage[export]{adjustbox} 
\usepackage{mathtools, cuted}
\usepackage{natbib}
\usepackage{enumerate}
\usepackage{enumitem}

\usepackage{algorithm}
\usepackage{algorithmic}

\usepackage{kpfonts}
\DeclareMathAlphabet{\mathsf}{OT1}{cmss}{m}{n}

\SetMathAlphabet{\mathsf}{bold}{OT1}{cmss}{bx}{n}

\providecommand{\norm}[1]{\|#1\|}

\newcommand{\commentout}[1]{}

\newtheorem*{theorem*}{Theorem}

\title{\bf Differentially Private Estimation of Hawkes Process}

\author{
Simiao Zuo$^\dagger$\thanks{Working paper. Correspondence to \texttt{simiaozuo@gatech.edu}.}, \
Tianyi Liu$^\dagger$, \ Tuo Zhao$^\dagger$, \ Hongyuan Zha$^\diamond$ \\
$^\dagger$Georgia Institute of Technology \\
$^\diamond$The Chinese University of Hong Kong, Shenzhen
}
\date{}

\begin{document}

\maketitle

\begin{abstract}
\noindent
Point process models are of great importance in real world applications. In certain critical applications, estimation of point process models involves large amounts of sensitive personal data from users. Privacy concerns naturally arise which have not been addressed in the existing literature. To bridge this glaring gap, we propose the first general differentially private estimation procedure for point process models. Specifically, we take the Hawkes process as an example, and introduce a rigorous definition of differential privacy for event stream data based on a discretized representation of the Hawkes process. We then propose two differentially private optimization algorithms, which can efficiently estimate Hawkes process models with the desired privacy and utility guarantees under two different settings. Experiments are provided to back up our theoretical analysis.
\end{abstract}

\input{0-introduction}

\input{0-background}
\input{0-hawkes}

\input{0-DP-review}
\input{0-DP-bounded}

\input{0-DP-low-rank}
\input{0-experiment}
\input{0-conclusion}

\clearpage
\bibliography{main}
\bibliographystyle{ims}

\clearpage
\appendix
\input{0-appendix}

\end{document}

%% file: 0-introduction.tex
\section{Introduction}

Point process is a powerful tool to model event sequence data, which commonly exist in many real world applications, e.g., social platforms~\citep{yang2011like}, personalized healthcare~\citep{wang2018supervised} and financial transactions~\citep{bacry2015hawkes}. 
For example, Hawkes process~\citep{hawkes1971spectra} has been successfully applied to applications, such as social network analysis~\citep{yang2013mixture}, criminology~\citep{mohler2011self} and seismology~\citep{ogata1988statistical}, where the events exhibit a self-exciting structure.
Estimation of point process models in some of the aforementioned applications often involves large amounts of sensitive personal data (e.g., credit card transactions and medical records), such that privacy concerns naturally arise. Unfortunately, existing literature on estimation of point process models completely overlooks privacy protection issues.

Differentially private learning~\citep{dwork2006calibrating} is a common approach to address privacy. It usually injects random noise into the released results computed from the sensitive private data. The injected noise has two proprieties.
First, it is carefully controlled such that the noisy results are insensitive to a single change in the original dataset. This indicates that an input data point can only have limited impact on model estimation. Moreover, such a property ensures that an attacker cannot infer any information about a particular entry in the dataset, even if she/he possesses the remaining data.
Second, the estimated model under the injected noise is close to the model estimated from unperturbed inputs. This property guarantees that the results are useful in practice.
We highlight that the paper is not about preserving \emph{data privacy}, but rather estimating models in a differentially private way, which results in a \emph{differentially private model} (i.e., an attacker cannot infer data from it).

One major challenge of applying the differential privacy technique is to properly define privacy on event sequence data. This is because the analysis of privacy relies on the concept of ``neighboring datasets''~\citep{dwork2006calibrating}. Intuitively, the concept describes two datasets that only differ in one entry, and the distance (usually described by $\ell_1$) between these datasets needs to be bounded by a constant.
Such a scenario is implementable for discrete inputs, e.g., counts of instances. However, point processes are defined over continuous time, and the time gap between any two events can be arbitrarily large.

To define privacy on event sequence data, we propose to discretize the continuous input space into discrete bins, where each bin corresponds to a time interval. Correspondingly, the origin input data (given by time stamps) are transformed into bin counts.
The bin-count sequence can be effectively approximated by an integer-valued autoregressive model (INAR, \citealt{jin1991integer}).
Using such a discretization method, neighboring datasets are naturally defined as two point processes whose corresponding bin-count sequences only differ in one entry by one count.

We remark that discretization only partially protects the time stamps. This is because in practice the bin size is usually application-specific, and a small bin size cannot provide enough protection.
Also, discretization is not enough to protect privacy. For example, consider a multi-variate point process, where each variate corresponds to a social network and each event is a tweet. After discretization, even though the exact time stamps are not observable, the attacker still knows the number of events in each bin (time interval), and thus can deduce critical information such as user preference. As another example, in practice it is likely that no event happens within a time interval, e.g., no tweet from a user after midnight. However, without protecting the bin-count sequence, an attacker can infer sensitive information such as a user's schedule.
Therefore, we provide privacy protection to the bin-count sequence. Intuitively, this means that the number of events within a time interval is confounded, i.e., the attacker do not know whether any event happens within a time interval, or exactly how many events happened.

This paper uses the multi-variate Hawkes process as an example, and proposes two private algorithms to estimate point process models under two different settings, respectively. The first setting assumes the intensity of the multi-variate Hawkes process model has a bounded Frobenius norm when discretized into a matrix. Then we develop an efficient differentially private projected gradient descent algorithm for such a setting. 
The second setting assumes that the intensity of the multi-variate Hawkes process model has a low rank structure when discretized~\citep{bacry2015sparse, zhou2013learning, sheen2020tensor}.
Such an assumption stems from, e.g., community structures in social networks~\citep{zhou2013learning} and structured spatio-temporal interactions in criminology~\citep{sheen2020tensor}.
Then we develop an efficient differentially private conditional gradient algorithm for such a setting.
Theoretically, we establish privacy and utility guarantees for both the algorithms.

We remark that in addition to the natural definition of privacy, the adopted discretization approach has two advantages over conventional maximum likelihood estimation (MLE) methods.
First, estimation of the INAR model solves a convex optimization problem, which is numerically stable and global optimality can be guaranteed~\citep{kirchner2018nonparametric}. In contrast, the conventional MLE methods involve non-convex optimization and are often numerically unstable.
Second, the INAR model is more adaptive. Conventional MLE methods often need to specify a fixed kernel, e.g., an exponential kernel, to describe the interactions among events. Such a fixed kernel makes over-simplified assumptions of the complicated interaction dynamics, which limits the expressive power of point process models. For example, the exponential kernel cannot model the case where the interactions change periodically.

To the best of our knowledge, we are the first to apply differential privacy to point process model estimation. We summarize our contributions as the following: 
(1) We rigorously define differential privacy for event sequence data; (2) We propose two differentially private algorithms with privacy and utility guarantees.

The remainder of this paper is organized as follows: In Section~\ref{sec:background} we introduce background information on multi-variate Hawkes process and differential privacy. In Section~\ref{sec:diff-definition} we formally define privacy on a discretized representation of point processes. In Section~\ref{sec:algorithms} we propose a projected gradient descent and a condition gradient algorithm with privacy and utility guarantees. Section~\ref{sec:experiments} contains numerical experiments. We conclude and discuss this paper in Section~\ref{sec:conclusion}.

\vspace{0.1in}
{\bf Notations:}
Denote $[n]=\{1,2,\cdots,n\}$. Let $\cS$ be a subspace of $\RR^d,$ and we use $\mathrm{Proj}_{\cS}(\cdot)$ to denote the projection of a vector or a matrix to $\cS$. Let $\boldsymbol{1}_{d_1\times d_2}$ denotes a matrix with all entries being 1. For matrices $A,B\in \RR^{n\times m}$, $\langle A, B\rangle$ denotes the Frobenius inner product, i.e., $\langle A, B\rangle=\tr(A^\top B)$. Let $\norm{A}_{\rm{F}}$ and $\norm{A}_*$ denote the Frobenius norm and the nuclear norm of a matrix $A$, respectively.

%% file: 0-background.tex
\section{Background}
\label{sec:background}

\noindent $\diamond$
\textbf{Hawkes Process}~\citep{hawkes1971spectra} is a doubly stochastic point process.
For a $M$-variate Hawkes process, the intensity function for any $m \in [M]$ is
\begin{align} \label{eq:inten-hp}
    \lambda_m(t) = \mu_m + \sum_{i=1}^M \sum_{j: t_j<t} \psi_i (t-t_j).
\end{align}
Here $\psi(\cdot)$ is the kernel function that captures interactions among events, and $\mu_m$ is the base intensity, which describes the arrival of events triggered by external sources, i.e., events that are independent to other events in the process.
An common approach is to specify $\psi(\cdot)$ in \eqref{eq:inten-hp} as a decaying function, e.g., exponential function and power-law function.
Intuitively, with such a choice of kernel, the intensity function means that occurrence of the current event is positively influenced by all the past events, and the influence decays through time.

As previously mentioned, the major drawback of the formulation in \eqref{eq:inten-hp} is that it makes over-simplified assumptions of the interaction dynamics.
Recently, adaptive estimation procedures of Hawkes process were proposed~\citep{kirchner2016hawkes, kirchner2017estimation, kirchner2018nonparametric}. These models are more flexible to capture the underlying interaction dynamics. Another line of works use neural networks to parameterize the intensity function~\citep{du2016recurrent, mei2016neural, zuo2020transformer}, which provides even more flexibility.

Conventionally, the Hawkes process model \eqref{eq:inten-hp} is estimated by maximum-likelihood estimation. However, optimizing the likelihood is numerically challenging. An alternative is to adopt a least squares formulation for model estimation~\citep{hansen2015lasso, bacry2015sparse, kirchner2016hawkes, kirchner2017estimation, kirchner2018nonparametric}.

\vspace{0.1in}
\noindent $\diamond$
\textbf{Differential privacy}~\citep{dwork2006our, dwork2006calibrating} concerns whether the output of computation over a dataset can revel information about private data.
Various algorithms are proposed that add noise to the computation steps, e.g., the Laplace mechanism~\citep{dwork2006calibrating} and the exponential mechanism. The noise is carefully controlled such that privacy is preserved, and the computation result under the introduced noise is close to the noiseless version. See \citet{dwork2014algorithmic} for a detailed review.
Subsequently, differentially private updates such as gradient descent~\citep{song2013stochastic, bassily2014private, abadi2016deep} are proposed.
We remark that there are other works on point process that adopt different definitions of privacy \citep{walder2020privacy, mohler2018privacy}, whereas we focus on differential privacy. Therefore, these works are orthogonal to our focus.

%% file: 0-hawkes.tex
\section{Differential Privacy for Hawkes Process}
\label{sec:diff-definition}

The major challenge to apply differential privacy is that point processes are defined over continuous time, such that the time interval between any two events can be arbitrarily large. Therefore, before deriving differentially private algorithms, we have to properly define privacy on event sequence data.

Let $(\Omega, \PP, \cF)$ be a probability space.
On this space, we have stochastic point sets $\cP^{(1)},..., \cP^{(d)}\subset \RR$ of the form $\cP^{(i)} = \{T^{(i)}_1,T^{(i)}_2,\cdots\}, \forall i\in [d]$. Note that we consider finite time events, i.e., $t \leq T, \forall t \in \cP^{(i)}, \forall i \in [d]$.
Denote $\Nb$ a $d$-variate counting process corresponding to the point sets, where $\Nb=(N^{(1)},\cdots, N^{(d)})^\top$. Here $N^{(i)}(t)$ counts the number of points in $\cP^{(i)}$ before time $t$.

For a multi-variate Hawkes process, we define its base intensity $\eta \in \RR_{\ge 0}^d$ and its excitement function $H=(h_{i,j})_{1\le i,j \le d}: \RR_{\ge 0} \rightarrow \RR_{\ge 0}^{d\times d}$, which is a matrix-valued function.
The conditional intensity of the Hawkes process is then
\begin{align} \label{eq:hawkes}
    {\Lambda}_{\bN}(t)
    = \lim_{\delta \downarrow 0} \frac{\EE \left[ N(t,t+\delta) | \cH_t^{\bN} \right]}{\delta}
    = \eta + \int_{-\infty}^t H(t-s) \ N(ds),
\end{align}
where
\begin{align*}
    \left( \int_{-\infty}^t H(t-s) \ N(ds) \right)_i
    = \left( \sum_{j=1}^d \int_{-\infty}^t h_{i,j}(t-s) \ N^{(j)}(ds) \right)_i.
\end{align*}
Here the filtration $\cH_t^{\bN}=\sigma\left( \left\{ \omega \in \Omega: \Nb\left((a,b]\right)=n \right\}, n \in \NN_0^d, a<b\leq t \right)$ is the history of the Hawkes process.  

We consider samples in finite time $(0,T]$.
For some bin size $\Delta > 0$, we construct a $\NN_0^d$-valued bin-count sequence as the following:
\begin{align} \label{eq:bin}
    X_{k} = \left( N^{(j)} \left( (k-1)\Delta, k\Delta \right) \right)_{j=1,\cdots,d},
    \quad k = 1,\cdots, n := \left\lfloor T/\Delta \right\rfloor.
\end{align}
Note that we have $n$ bins, and $X_k \in \RR^d$. We further denote $X = \{X_k\}_{k=1}^n$ the discretized data.

We remark that even though the discretization in \eqref{eq:bin} partially protects the time stamps, the bin counts still contain sensitive information and need private protection. For example, by comparing user activities (i.e., the number of events) within a time interval, an attacker can infer critical behavioral information. To resolve this issue, we add privacy protections to the bin-count sequences.

We can now define differential privacy for Hawkes process, given the discretized representation $X$. As in the standard definition of differential privacy~\citep{dwork2006calibrating}, we need to first clarify the distance between two datasets.

\begin{definition}[Distance between Datasets]
The distance between two datasets $D =\{D_i\}_{i=1}^n$ and $D'=\{D_i'\}_{i=1}^n$ where $D_i, D_i' \in \RR^d$ are the data is defined as follows:
$$\cD(D,D') = \sum_{i=1}^n \norm{D_j-D_j'}_1,$$
where $\norm{\cdot}_1$ is the $\ell_1$ norm. Moreover, we call $D$ and $D'$ a pair of neighboring datasets if
$$\cD(D, D')\leq 1.$$
\end{definition}

Since the discretized bin-count sequence $X$ takes integer values, two neighboring datasets can only vary in one bin by one count. We are now ready to formally define differential privacy.

\begin{definition}[Differential Privacy for Event Sequence Data]\label{def:privacy}
A randomized algorithm $\cA$ is ($\epsilon$, $\delta$)-differentially private if for all $S \in \text{Range}(\cA)$ and neighboring datasets $D$, $D'$:
$$ \PP[\cA(D) \in S] \leq e^\epsilon \PP[\cA(D') \in S] + \delta. $$
\end{definition}

The algorithm $\cA$ is our Hawkes process estimation procedure. Intuitively, a randomized algorithm that achieves differential privacy will behave similarly on similar input bin-count sequences.

%% file: 0-DP-review.tex
\section{Differentially Private Estimation of Hawkes Process}
\label{sec:algorithms}

In this section, we first review a least squares estimator for the base intensity and the excitement function \eqref{eq:hawkes}.
Then, we propose two estimation procedures that preserve privacy. 
All the proofs are deferred to the appendix.

\subsection{An Estimation Procedure of Hawkes Process}

Our goal is to estimate the base intensity and the excitement function defined in \eqref{eq:hawkes}.
Based on the discretized representation \eqref{eq:bin}, a least squares estimator~\citep{kirchner2017estimation} can be constructed using the INAR($p$) model~\citep{jin1991integer}, which is an integer-valued autoregressive model of order $p$.
Concretely, for some support $s$, where $0<\Delta < s < T$, we define the maximal lag $p=\lceil s/\Delta \rceil$. Then the conditional least squares estimator $\theta_\textsubscript{CLS} \in \RR^{d\times (dp+1)}$ solves
\begin{align*}
    & \min_\theta \norm{\theta Z Z^\top - Y Z^\top}_{\rm F}^2,
\end{align*}
where $Z$ is the design matrix defined as
\begin{align*}
    & Z(X) :=
    \begin{bmatrix}
        X_p & X_{p+1} & \cdots & X_{n-1} \\
        X_{p-1} & X_{p} & \cdots & X_{n-2} \\
        \cdots & \cdots & \cdots & \cdots \\
        X_1 & X_{2} & \cdots & X_{n-p} \\
        1 & 1 & \cdots & 1
    \end{bmatrix}
    \in \RR^{(dp+1) \times (n-p)}
\end{align*}
and $Y(X) = \left[ X_{p+1}, X_{p+2}, \cdots, X_{n} \right] \in \RR^{d \times (n-p)}$.

From $\theta\textsubscript{CLS}$, the base intensity and the excitement function is 
\begin{align*}
    & \left[ H_1, \cdots, H_p, \eta \right] \
    := H \
    := \frac{1}{\Delta} \theta_\textsubscript{CLS}.
\end{align*}

Note that $H \in \RR^{d\times (dp+1)}$, $H_i \in \RR^{d\times d}$ and $\eta \in \RR^d$.
Through this estimation procedure, we acquire a sequence of estimators for the excitement function $\{H_i\}_{i=1,\cdots,p}$ over a grid. The support of the grid is $\{\Delta, 2\Delta, \cdots, s=p\Delta\}$.
Here $H$ is a discretized version of the true excitement function \eqref{eq:hawkes}, which can be recovered by simply interpolate between the point-wise estimators $\{H_i\}$, e.g., using linear interpolation or cubic splines.

Under the assumption that $\frac{1}{n-p}ZZ^\top$ is invertible and converging~\citep{kirchner2017estimation}, we know for $0<t<s,$ $[H_{\lfloor t/\Delta \rfloor}]_{ij}$ and $[\eta]_i$ are weakly consistent estimators (when $T\rightarrow \infty$, $\delta\rightarrow 0$ and $s=\Delta p\rightarrow \infty$) for $h_{i,j}(t)$ and $\eta_i$, respectively.


\begin{assumption}\label{ass:bounded}
There exists a constant $R>0$, such that $\{X_k\}_{k=1,...,n}$ satisfies
\begin{align*}
\norm{\frac{1}{n-p}ZZ^\top}^2_\mathrm{F}\leq R\quad\textrm{and}\quad	\norm{\frac{1}{n-p}YZ^\top}^2_\mathrm{F}\leq R.
\end{align*}
\end{assumption}

Assumption \ref{ass:bounded} holds for sufficiently large $n$ when the Hawkes process is stationary and is strongly-mixed.
Existing literature has shown that under mild conditions, stationary Hawkes process satisfies strong mixing properties~\citep{cheysson2020strong,Boly2020mixing}.

%% file: 0-DP-bounded.tex
\subsection{Differentially Private Estimation of Bounded Parameters}
\label{sec:bounded}

We discuss the case where the parameter matrix $H$ is bounded. That is, there exists a constant $B>0$ such that $\norm{H}_\mathrm{F} \leq B$.
We consider the following formulation:
\begin{align}\label{eq:bounded}
\Delta H
    & = \argmin_{ U\in\cB}  \cL( U; Z(X),Y(X))
    = \argmin_{ U\in\cB} \frac{1}{2(n-p)^2}\norm{ U ZZ^\top  - YZ^\top}^2_\mathrm{F},
\end{align}
where $\cB = \left\{ U\in \RR^{d\times (dp+1)}\big| \norm{ U}_\mathrm{F}\leq \Delta B\right\}$, and the scalar $\frac{1}{2(n-p)^2}$ is added to avoid explosion of $ZZ^\top$.
Through the rest of the paper, if not clearly specified, we will simply write  $\cL(U; Z(X),Y(X))$ as $\cL(U; X)$. 

To solve \eqref{eq:bounded}, a common algorithm is projected gradient descent. Concretely, we take the following update at the $k$-th iteration:
\begin{align*}
  & U_{k+1} =\mathrm{Proj}_\cB \left(U_k - \eta  \nabla \cL(U_k;X)\right).
\end{align*}
Note that this requires access to the private data $X$.
To preserve privacy, we perturb the gradient with a Gaussian noise and then apply projected gradient descent as follows:
\begin{align*}
    U_{k+1} =\mathrm{Proj}_\cB \left( U_k - \eta\left(
    \nabla \cL(U_k;X) + \boldsymbol{\zeta}_k \right) \right),
\end{align*} 
where $\boldsymbol{\zeta}_k \sim \cN\left(0, \sigma^2\mathbf{1}_{d\times (dp+1)}\right)$ is drawn from a normal distribution.
Details of the algorithm can be found in Algorithm~\ref{alg:DP-GD}. 

\begin{algorithm}[tb!]
\caption{Differentially Private Estimation using Projected Gradient Descent.}
\label{alg:DP-GD}
\begin{algorithmic}
\STATE \textbf{Input: } $X$: data; $\cL(U;X)$: objective function; $\epsilon$: privacy budget; $p:$ time lag; $\Delta:$ bin size; $\sigma^2:$ noise variance; $\eta:$ learning rate.
\STATE \textbf{Initialize: } Choose an arbitrary $ U_0\in \cB$;
\FOR {$k = 1, \cdots, K-1$}{
	\STATE Sample $\boldsymbol{\zeta}_k\sim\cN\left(0, \sigma^2\mathbf{1}_{d\times (dp+1)}\right)$;
	\STATE $U_{k+1}' = U_k - \eta\left( \nabla \cL( U_k;X) +\boldsymbol{\zeta}_k\right)$;
	\STATE $U_{k+1} =\mathrm{Proj}_\cB ( U_{k+1}')$;
}
\ENDFOR
\STATE \textbf{Output: } $H^{\text{priv}} = \frac{1}{\Delta}  U_K$
\end{algorithmic}
\end{algorithm}
 
Theorem~\ref{thm:gd_privacy} shows that under a suitable noise level, Algorithm~\ref{alg:DP-GD} attains $(\epsilon, \delta)$-differential privacy.
\begin{theorem}[Privacy Guarantee] \label{thm:gd_privacy}
Under Assumption \ref{ass:bounded},  if we take $$\sigma^2= \frac{8(\Delta B)^2 R^2K\log^2\frac{K}{\delta}}{\epsilon^2},$$ Algorithm \ref{alg:DP-GD} is $(\epsilon,\delta)$-differentially private.
\end{theorem}

In the following theorem, we provide utility guarantee for Algorithm \ref{alg:DP-GD}.
\begin{theorem}[Utility Guarantee]\label{thm:gd_utility}
Under Assumption \ref{ass:bounded},  if we take
\begin{align*}
  & \sigma^2= \frac{8(\Delta B)^2 R^2K\log^2\frac{K}{\delta}}{\epsilon^2}, \quad
  \eta_k = \frac{\Delta B}{\sqrt{k(4\Delta B)^2 R^2 + d(dp+1)\sigma^2)}},  
\end{align*}
and the total number of iterations $K = \cO(1),$ then
\begin{align*}
   \EE[ \cL(U_{K};X) - \min_{U\in\cC}\cL(U,X)] 
   = \cO \left( \frac{(\Delta B)^2\sqrt{ d(dp+1)}}{\epsilon}\log\frac{1}{\delta} \right). 
\end{align*}
Here the expectation is over the randomness of the algorithm.
\end{theorem}

%% file: 0-DP-low-rank.tex
\subsection{Differentially Private Estimator of Low Rank Parameters}
\label{sec:low_rank}

In this section, we consider the case where the parameter matrix $H$ is low-rank. Such a scenario is common in real-life. For example, in social networks~\citep{zhou2013learning}, different variates can share the same confounders, such that the excitement matrix of the multi-variate Hawkes process is approximately low rank.
To promote such a low-rank structure, we add a constraint $\norm{H}_* \leq r$ to the least squares problem as the following:
\begin{align}\label{opt:low_rank}
\Delta H
    &= \argmin_{U\in \cC}  \cL(U; X)
    = \frac{1}{2(n-p)^2}\norm{U ZZ^\top  - YZ^\top}^2_\mathrm{F}, 
\end{align}
where $\cC = \left\{U\in \RR^{d\times (dp+1)}\big| \norm{U}_*\leq r\right\}.$ Note that $\cC$ is a convex set. Therefore, we can apply the conditional gradient method to solve this convex constraint problem. Concretely, at the $k$-th iteration, we take the following update:
\begin{align*}
    U_{k+1} = (1-\mu) U_k 
    +\mu \argmin_{U \in \cC} \left\langle U, \nabla \cL(U_k; X) \right\rangle.
\end{align*}

Similar to Algorithm \ref{alg:DP-GD}, we perturb the gradient with a Gaussian noise to preserve privacy. The differentially private update rule is then
\begin{align*}
    U_{k+1} = (1-\mu) U_k +\mu\argmin_{U\in \cC}\left\langle U, \nabla \cL(U_k; X)+ \boldsymbol{\zeta}_k \right\rangle, 
\end{align*}
where $\boldsymbol{\zeta}_k \sim\cN \left(0, \sigma^2\mathbf{1}_{d\times (dp+1)}\right)$. See Algorithm~\ref{alg:DP-FW} for more details.

\begin{algorithm}[tb!]
\caption{Differentially Private Estimation using Conditional Gradient Method.}
\label{alg:DP-FW}
\begin{algorithmic}
\STATE \textbf{Input: }$X$: data; $\cL(U;X)$: objective function; $\epsilon$: privacy budget;  $p:$ time lag; $\Delta:$ bin size;  $\sigma^2:$ noise variance.
\STATE \textbf{Initialize: } Choose an arbitrary $U_0\in \cC$;
\FOR{$k = 1, \cdots, K-1$}{
    \STATE Sample $\boldsymbol{\zeta}_k \sim\cN \left(0,  \sigma^2\mathbf{1}_{d\times (dp+1)}\right)$;
    \STATE Set $\alpha(U) = \left\langle U, \nabla  \cL(U_k; X)+ \boldsymbol{\zeta}_k \right\rangle$;
  	\STATE Set $\tilde U_k = \argmin_{U\in \cC}\alpha(U)$;
    \STATE Let $U_{k+1} = (1-\mu) U_k +\mu \tilde U_k, $ where $\mu=\frac{1}{K+2}$;
}
\ENDFOR
\STATE \textbf{Output: } $H^{\text{priv}} = \frac{1}{\Delta} U_K$
\end{algorithmic}
\end{algorithm}

The following theorem states that Algorithm \ref{alg:DP-FW} is $(\epsilon,\delta)$-differentially private given a properly chosen noise level $\sigma^2$.
\begin{theorem}[Privacy Guarantee]\label{thm:privacy_FW}
Under Assumption \ref{ass:bounded},  if we take
$$\sigma^2 =  \frac{128 K\Delta^2 r^2 R^2\log^2\frac{K}{\delta}}{\epsilon^2},$$
Algorithm \ref{alg:DP-FW} is $(\epsilon,\delta)$-differentially private.
\end{theorem}

Before providing utility guarantee for the private conditional gradient method, we first state the definition of curvature constant and Gaussian width.

\begin{definition}[Curvature Constant]
For $f: \bX\rightarrow \RR,$ we define the curvature constant $\Gamma_f$ as follows.
\begin{align*}
  \Gamma_f = \sup_{x_1,x_2\in\bX, \gamma \in(0,1]} ~
  \frac{2}{\gamma^2} \Big( f((1-\gamma)x_1+\gamma x_2)
  - f(x_1) -\langle\gamma(x_2-x_1), \nabla f(x_1)\rangle \Big).
\end{align*}
\end{definition}
Let $\Gamma_{\cL}$ be the curvature constant of $\cL.$ Since $\cC$ is centrally symmetric, we have the following \citep{talwar2015nearly}
\begin{align*}
    \Gamma_{\cL} \leq 4\max_{U\in \cC} \norm{\frac{1}{n-p} UZZ^\top}_{\rm F}^2 \leq 4(\Delta r)^2R.
\end{align*}

\begin{definition}[Gaussian Width]\label{def:gaussian_width}
Let $B\sim \cN(0,\mathbf{1}_{d_1\times d_2})$ be a Gaussian random matrix in $\RR^{d_1\times d_2}.$	The Gaussian width of a set $\cS$ is defines as $\omega(\cS) = \EE_B\left[\sup_{s\in\cS}|\langle B,s\rangle|\right].$
\end{definition}

With the above definitions, we have the following utility guarantee.
\begin{theorem}[Utility Guarantee]\label{thm:utility_FW}
Under Assumption \ref{ass:bounded}, if we take
\begin{align*}
    & \sigma^2 = \cO\left(\frac{K\Delta^2 r^2 R^2\log^2\frac{K}{\delta}}{\epsilon^2}\right), \quad
    K =\cO\left(\left(\frac{\Delta r\epsilon}{\sqrt{dp+1}}\right) ^{\frac{2}{3}}\right),
\end{align*}
then we have the following utility guarantee:
\begin{align*}
\EE[ \cL(U_{K};X) - \min_{U\in\cC}\cL(U,X)]
=\cO\left(
\frac{R(\Delta r)^{\frac{4}{3}}(dp+1)^{\frac{1}{3}}{\log(\frac{1}{\delta})}}{\epsilon^{\frac{2}{3}} }
\right).   
\end{align*}
Here the expectation is taken over the randomness of the algorithm.
\end{theorem}


%% file: 0-experiment.tex
\section{Experiments}
\label{sec:experiments}

\begin{figure*}[t!]
    \centering
    \begin{subfigure}{0.32\textwidth}
        \centering
        \includegraphics[width=1.0\textwidth]{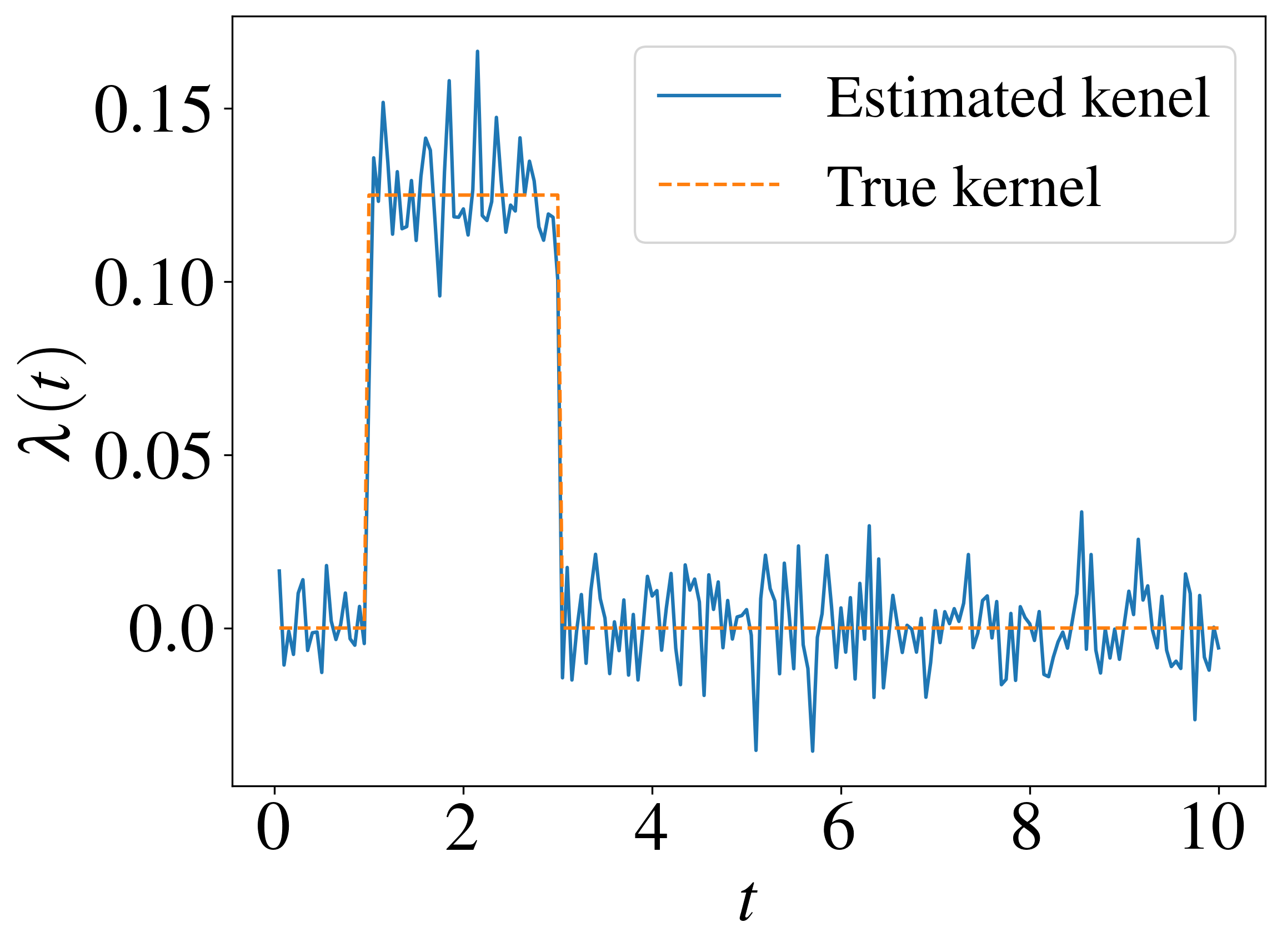}
        \caption{Non-private ($\sigma^2=0$).}
        \label{fig:kernel:nonprivate}
    \end{subfigure}%
    \begin{subfigure}{0.32\textwidth}
        \centering
        \includegraphics[width=1.0\textwidth]{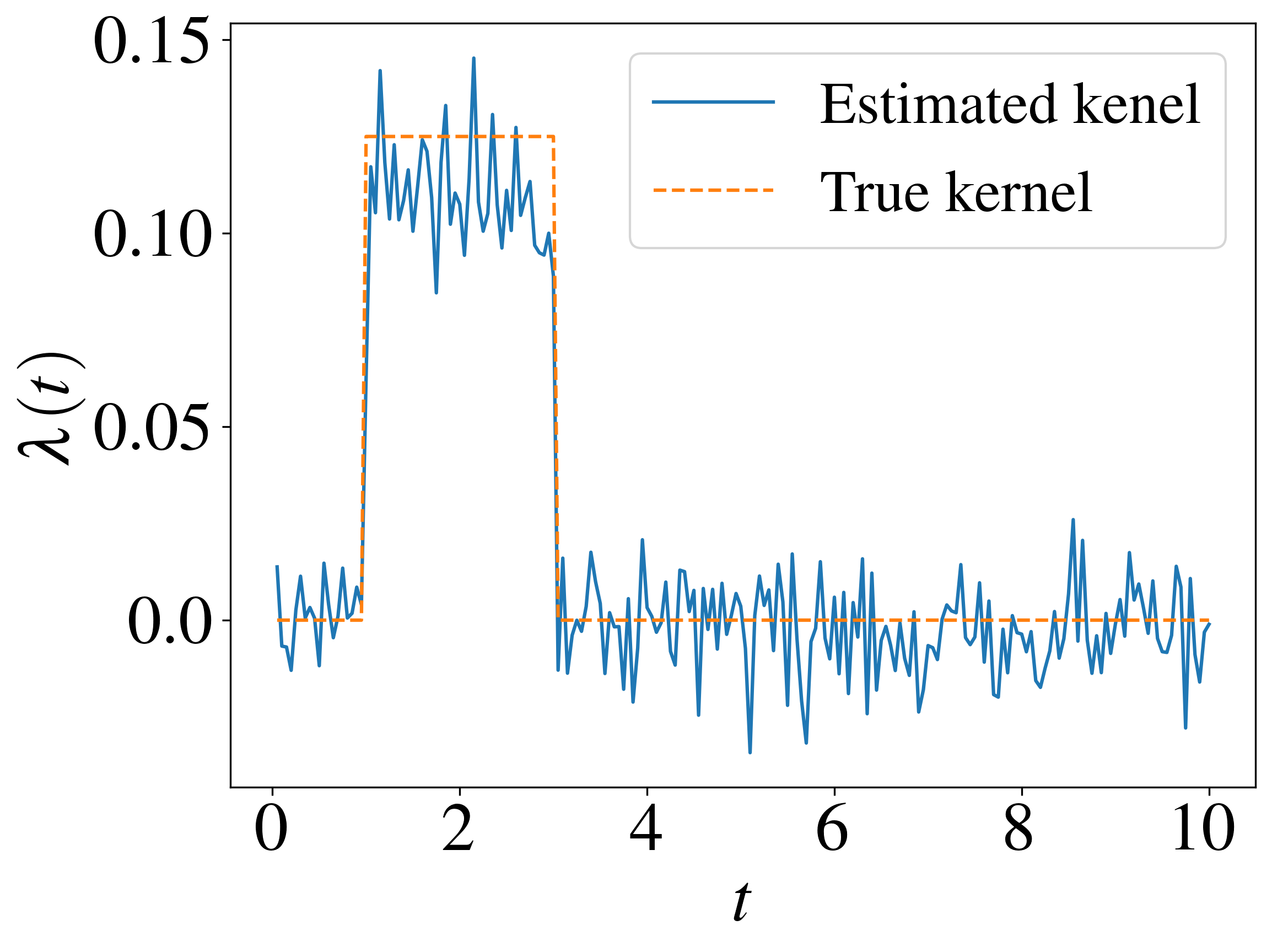}
        \caption{Private ($\sigma^2=0.01$).}
        \label{fig:kernel:small}
    \end{subfigure}%
    \begin{subfigure}{0.32\textwidth}
        \centering
        \includegraphics[width=1.0\textwidth]{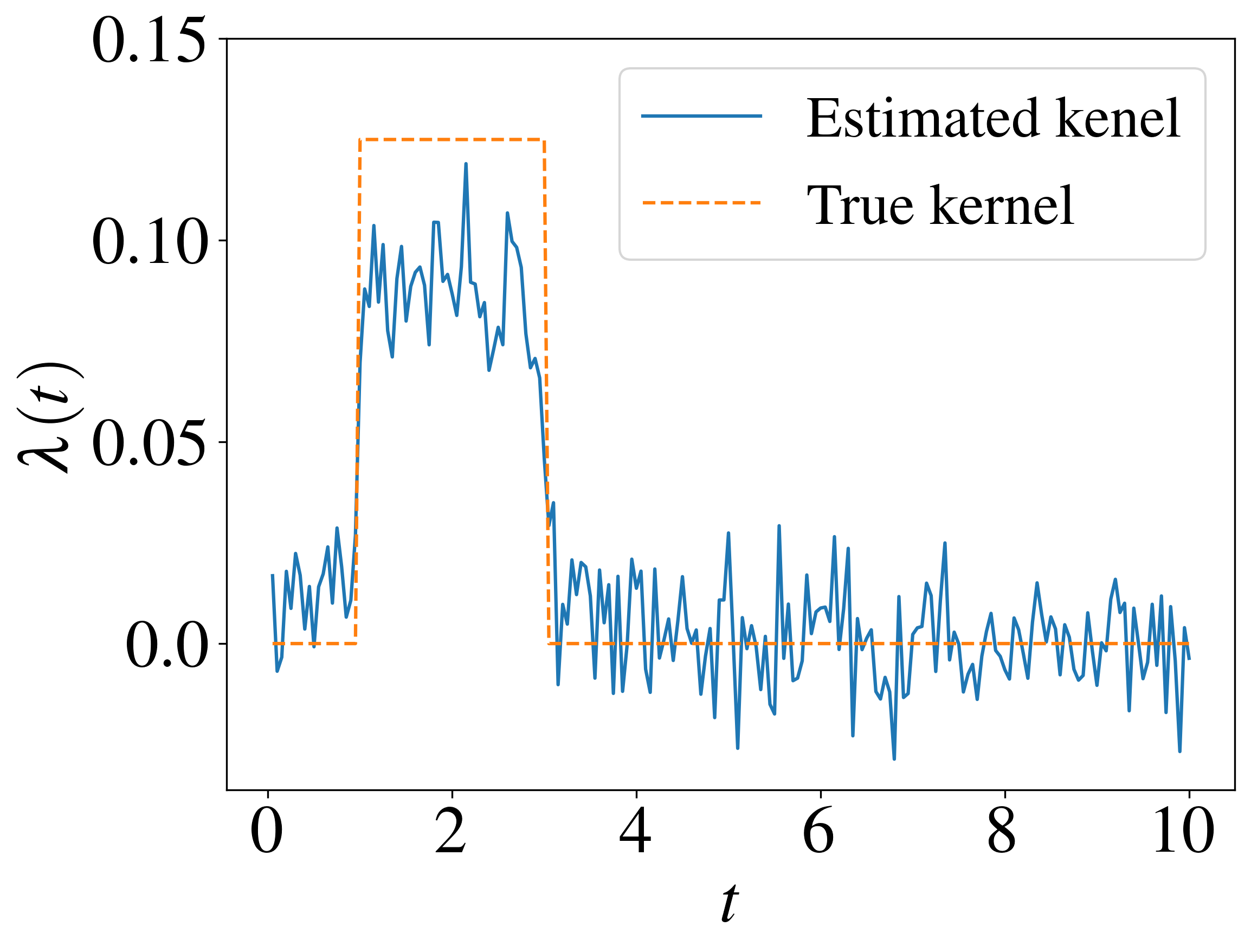}
        \caption{Private ($\sigma^2=10$).}
        \label{fig:kernel:large}
    \end{subfigure}%
    \caption{Non-private and private estimation of a kernel function $h_{1,2}$ under different noise level. Models are estimated using differentially private projected gradient descent.}
    \label{fig:kernel}
\end{figure*}

\begin{figure*}[t!]
    \centering
    \begin{subfigure}{0.34\textwidth}
        \centering
        \includegraphics[width=1.0\textwidth]{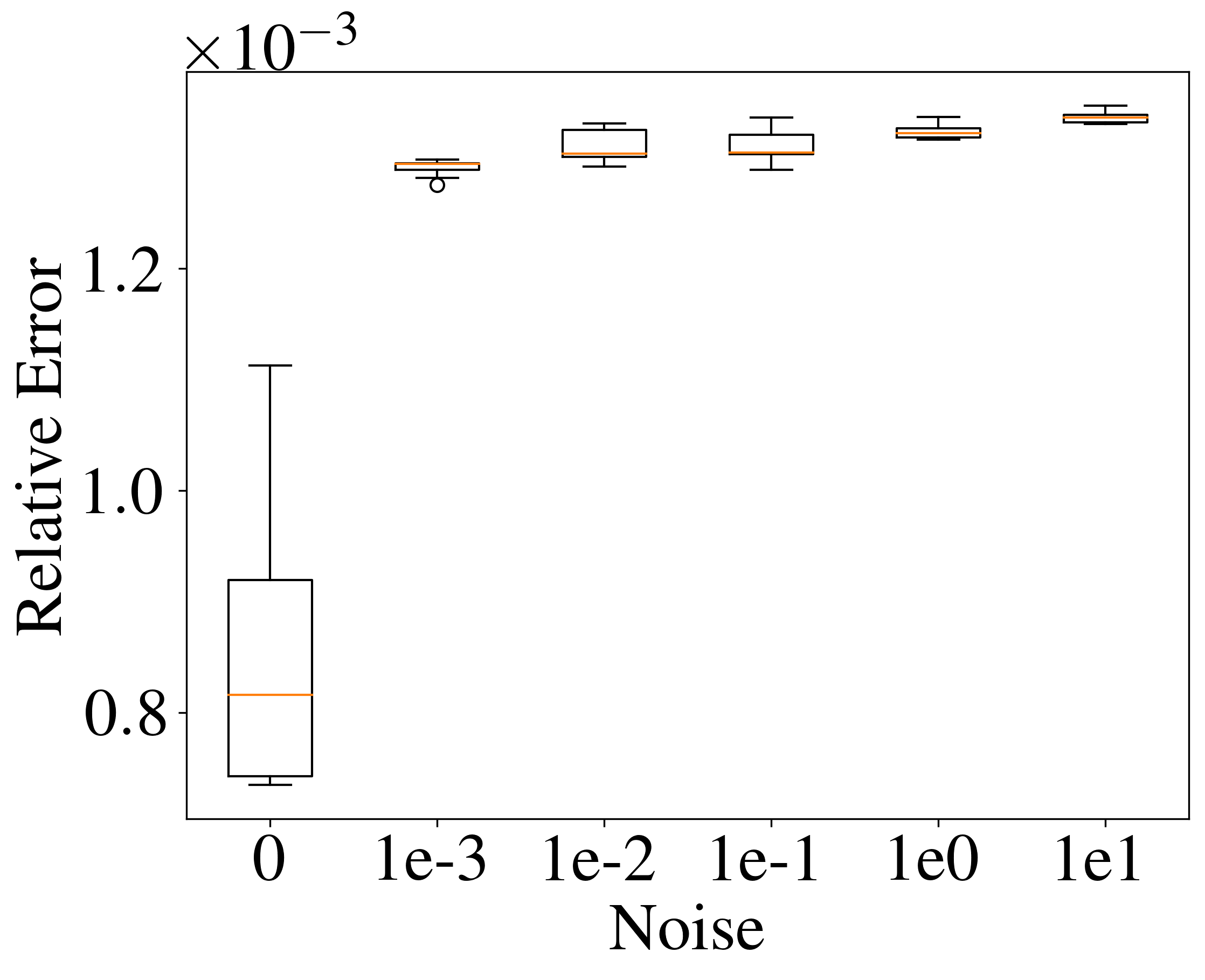}
        \caption{$\Delta=0.01$.}
    \end{subfigure}%
    \begin{subfigure}{0.31\textwidth}
        \centering
        \includegraphics[width=1.0\textwidth]{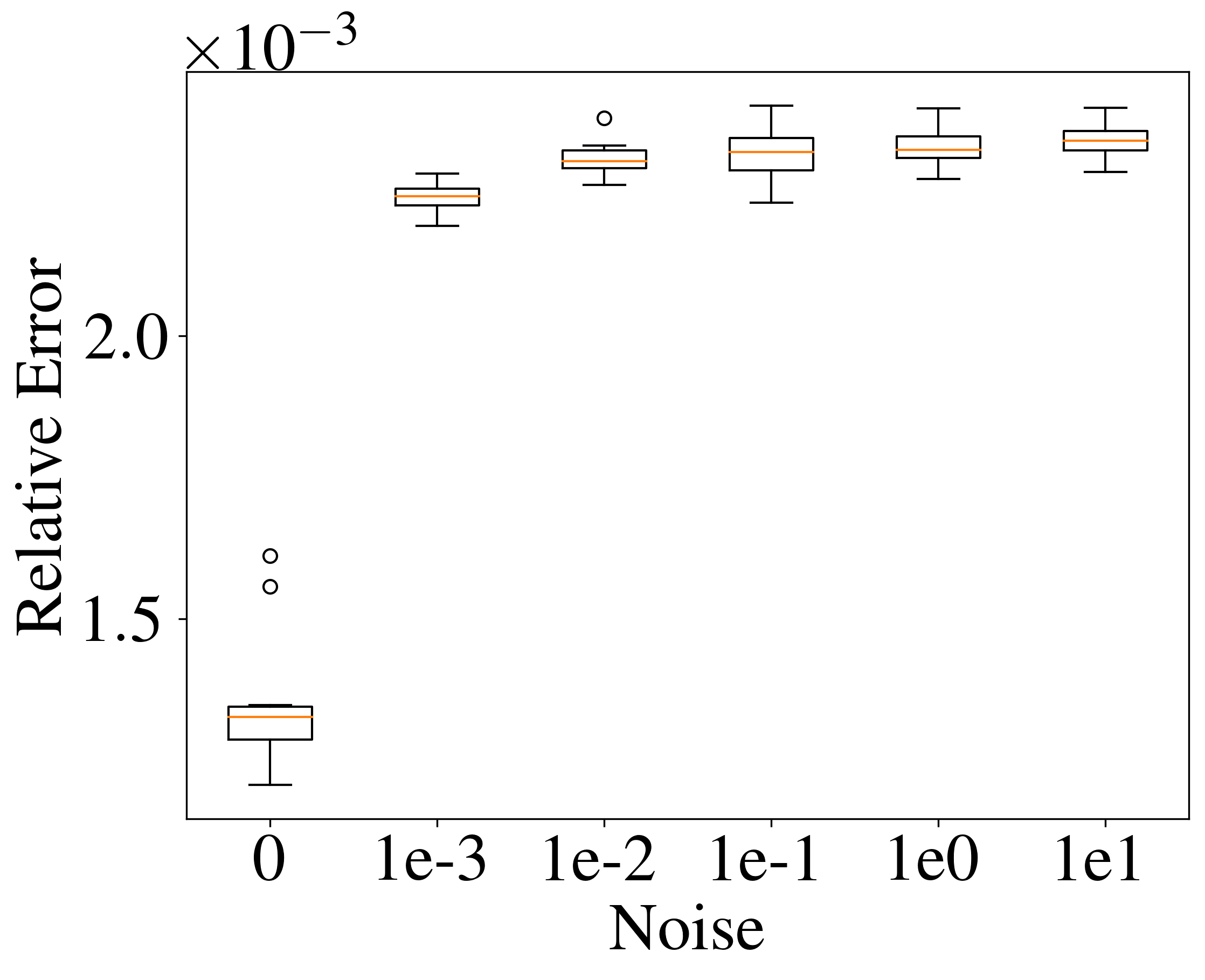}
        \caption{$\Delta=0.03$.}
    \end{subfigure}%
    \begin{subfigure}{0.31\textwidth}
        \centering
        \includegraphics[width=1.0\textwidth]{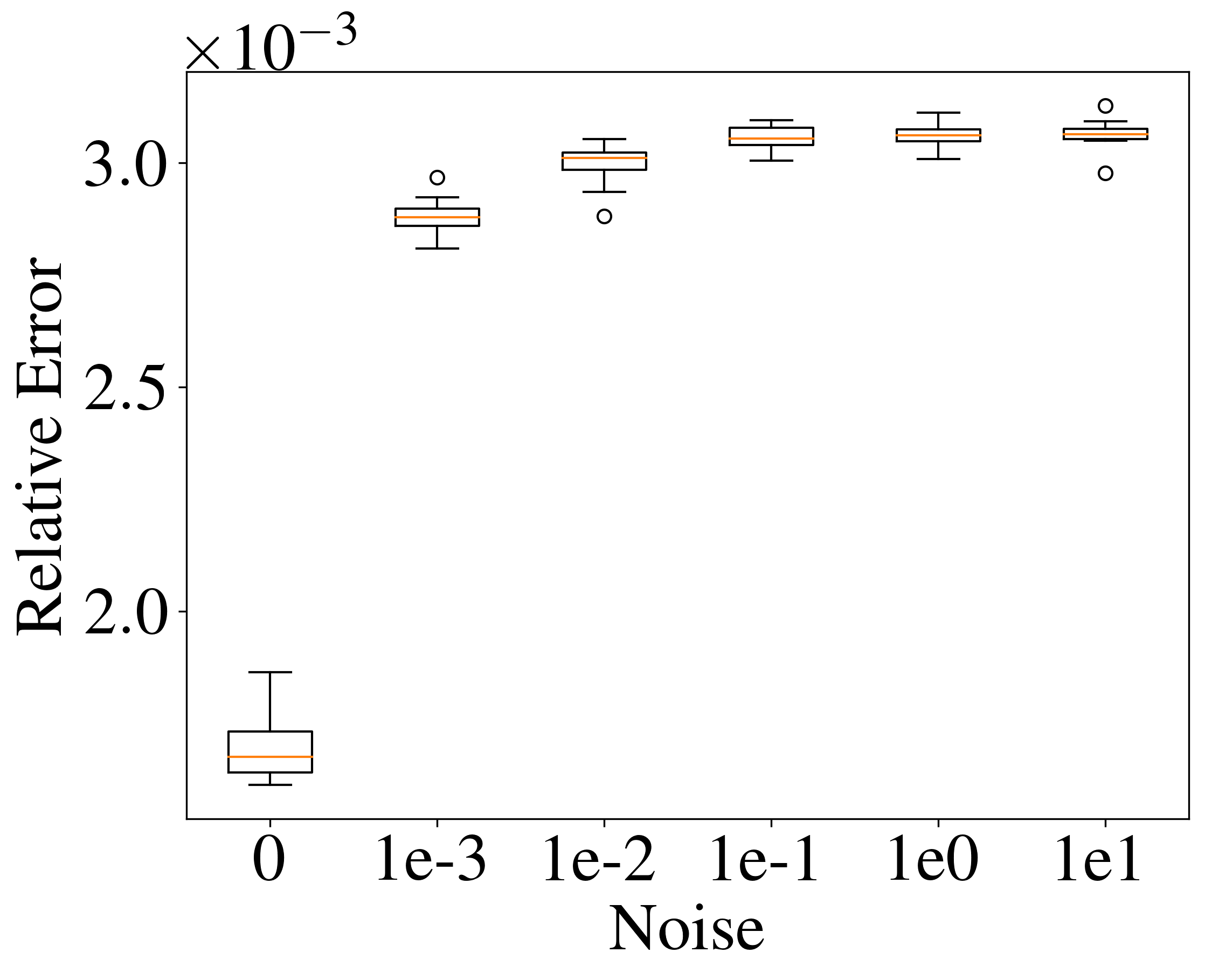}
        \caption{$\Delta=0.5$.}
    \end{subfigure}%
    \caption{Relative error of differentially private estimation under different noise level ($\sigma^2$) using different bin size. Each boxplot is produced from 10 runs. Here \textit{Noise=0} indicates a non-private estimation. Models are estimated using differentially private projected gradient descent.}
    \label{fig:bin}
\end{figure*}

We present experiments on synthetic and real-world datasets to study the proposed differentially private algorithms.
Specifically, we examine the relationship between the level of noise and estimation accuracy.
Note that the degree of attained privacy protection (i.e., $\epsilon$) can be calculated from the level of applied noise (i.e., $\sigma^2$).

\subsection{Differentially Private Projected Gradient Descent}
We consider a 2-variate Hawkes process, whose baseline intensity is $\eta=[0.25, 0.125]^\top$, and its excitement function is the following:
\begin{align} \label{eq:exp:intensity}
    H(t) &= 
    \begin{bmatrix}
        h_{1,1}(t) & h_{1,2}(t) \\
        h_{2,1}(t) & h_{2,2}(t)
    \end{bmatrix}
    = 
    \begin{bmatrix}
        0 & \mathbf{1}_{\{1 \leq t \leq 3\}} 0.125 \\
        \mathbf{1}_{\{2 \leq t \leq 4\}} 0.2 & 0.25 e^{-t}
    \end{bmatrix}.
\end{align}
Here $\mathbf{1}_{\{\cdot\}}$ is the indicator function. We use the \textit{tick} codebase~\citep{bacry2017tick} to simulate 1,000 events.
For differentially private estimation, we apply the proposed private projected gradient descent algorithm (Algorithm~\ref{alg:DP-GD}), where we set $B=0.2$ and $K=1000$.

Figure~\ref{fig:kernel} demonstrates the relationship between model estimation and noise level, where we visualize the estimation of $h_{1,2}$ as an example. Notice that for a non-private estimation (Fig.~\ref{fig:kernel:nonprivate}), the estimated intensity aligns almost perfectly with the ground truth intensity. When a small noise (i.e., $\sigma^2=0.01$ in Fig.~\ref{fig:kernel:small}) is applied, model estimation is still reliable. Moreover, the estimation does not collapse even under an excessive amount of noise (i.e., $\sigma^2=10$ in Fig.~\ref{fig:kernel:large}).

Figure~\ref{fig:bin} illustrates the relationship between estimation accuracy and bin size $\Delta$. Here relative error is measured as
\begin{align*}
    \text{Relative Error } =
    \frac{\norm{\widehat{H}- H\textsubscript{true}}_{\rm F}}{d (dp+1) \norm{H\textsubscript{true}}_{\rm F}},
\end{align*}
where $\widehat{H}$ is the model estimation and $H\textsubscript{true}$ is the ground truth discretized excitement function \eqref{eq:exp:intensity}. Note that we calculate entry-wise error to facilitate the comparison across different $\Delta$ (recall that $p$ inversely scales with $\Delta$).

From Figure~\ref{fig:bin}, we can see that no matter the choice of $\Delta$, the estimation accuracy does not drop significantly when applying the proposed differentially private estimation algorithm. For example, the relative error increase by less than 20\% even if we apply a very large noise ($\sigma^2=10$). Also note that by decreasing the bin size, estimation accuracy improves for both the non-private and the private estimation. We remark that for a fixed level of noise, decreasing the bin size improves privacy protection, i.e., $\epsilon$ increases when $\Delta$ decreases (see Theorem~\ref{thm:gd_privacy}).

\subsection{Differentially Private Conditional Gradient}
We present numerical experiments to study the proposed differentially private conditional gradient algorithm (Algorithm~\ref{alg:DP-FW}).

We consider a 4-variate Hawkes process, whose baseline intensity is
$$ \eta=[0.125, 0.125, 0.125, 0.125]^\top, $$
and its excitement function is 
\begin{align*}
    &H(t) = 
    \begin{bmatrix}
        H_0(t) & H_0^\top(t) \\
        H_0(t) & H_0^\top(t)
    \end{bmatrix}, \
    H_0(t) = 
    \begin{bmatrix}
        0 & h_{1,2}(t)\\
        h_{2,1}(t) & h_{2,2}(t)
    \end{bmatrix}, \\[3pt]
    &\text{where} \quad h_{12} = \mathbf{1}_{\{1 \leq t \leq 3\}} 0.125, \quad
    h_{21} = \mathbf{1}_{\{2 \leq t \leq 4\}} 0.25, \quad
    h_{22} = 0.2 e^{-t}.
\end{align*}
Notice that the discretized parameter matrix is low-rank (rank-2 with 4 variate).
We simulate 4000 events for this Hawkes process.
For differentially private estimation, we apply the proposed private conditional gradient algorithm (Algorithm~\ref{alg:DP-FW}), where we set $K=100$.

\begin{figure*}[t!]
    \centering
    \begin{subfigure}{0.32\textwidth}
        \centering
        \includegraphics[width=1.0\textwidth]{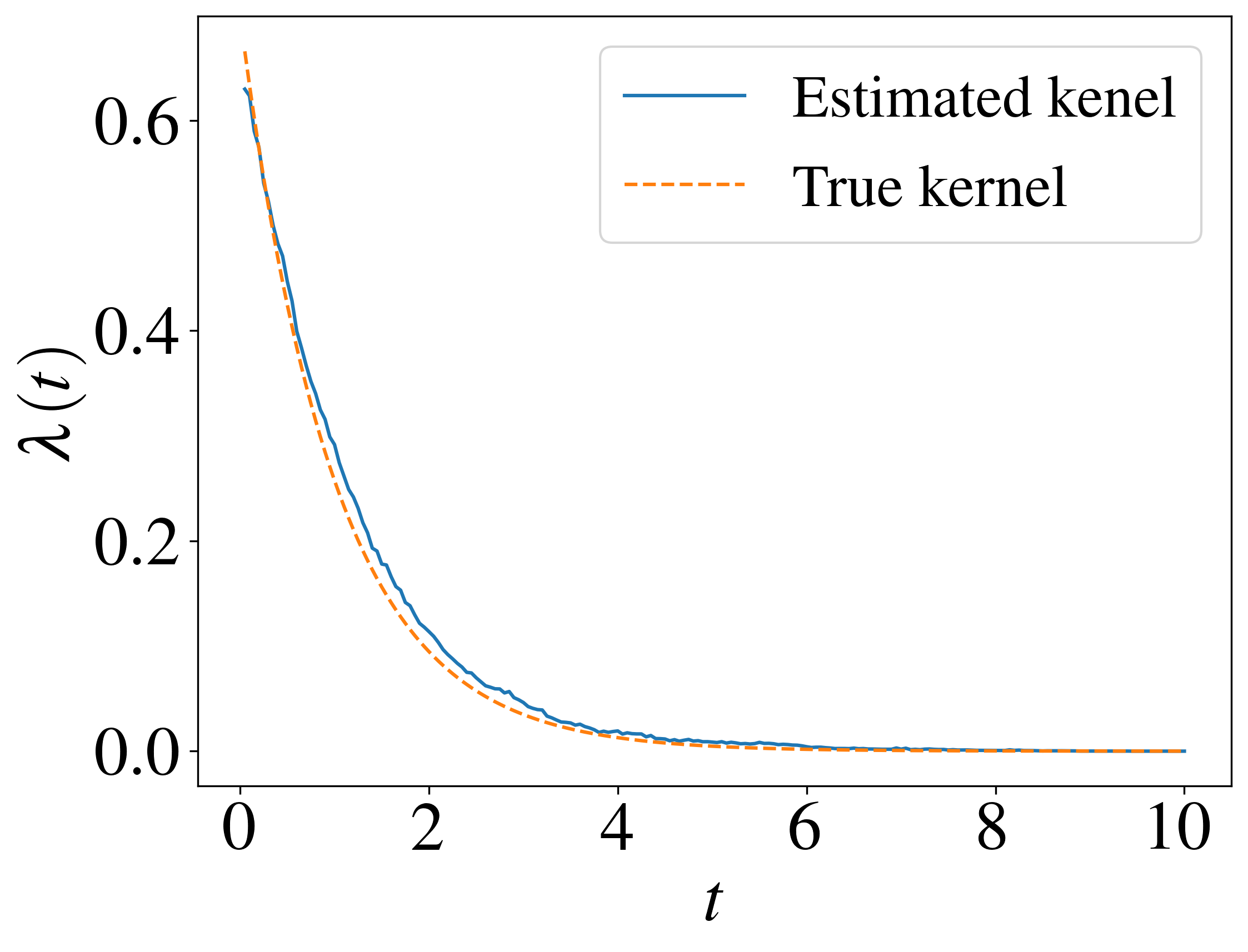}
        \caption{Non-private ($\sigma^2=0$).}
    \end{subfigure}%
    \begin{subfigure}{0.32\textwidth}
        \centering
        \includegraphics[width=1.0\textwidth]{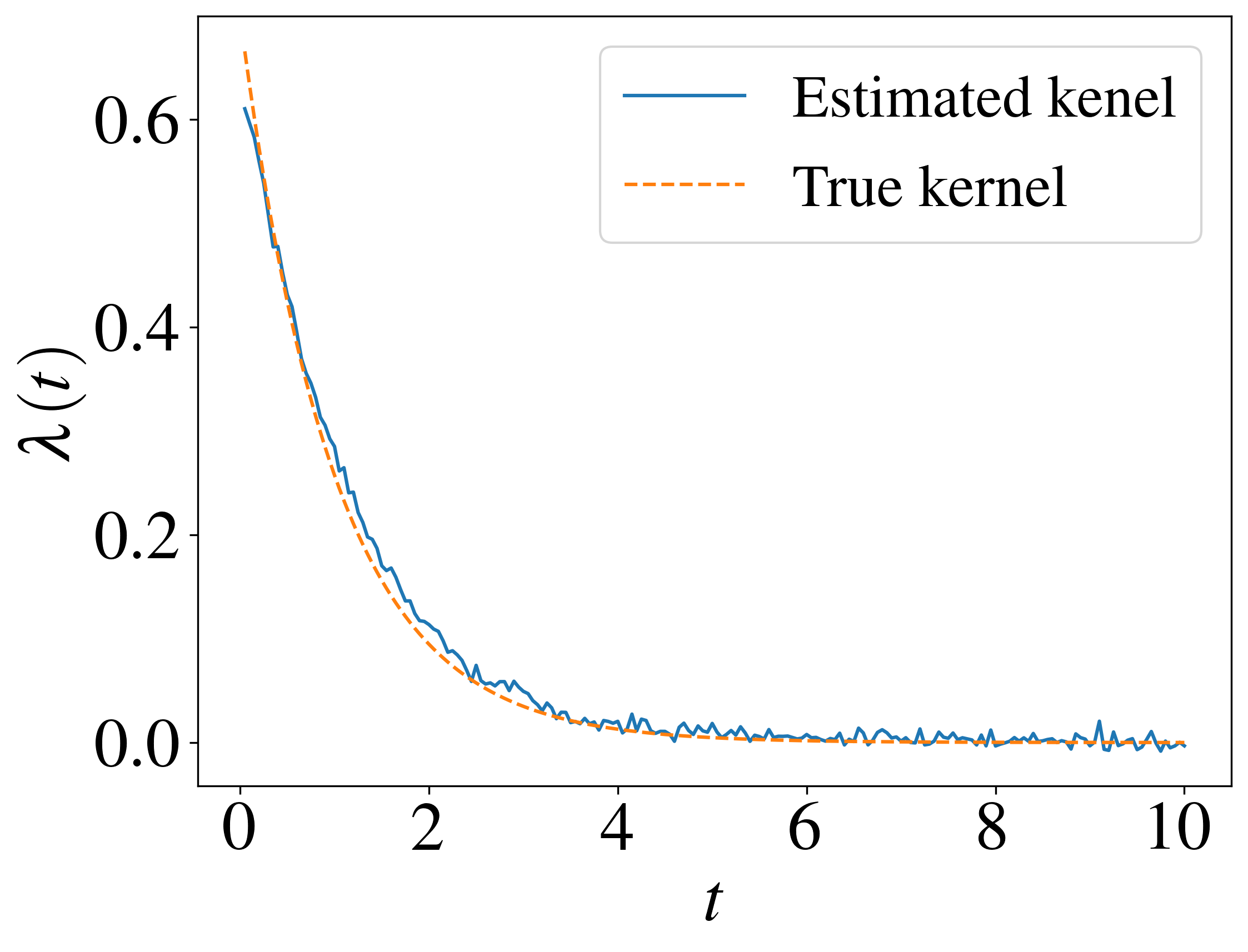}
        \caption{Private ($\sigma^2=0.01$).}
    \end{subfigure}%
    \begin{subfigure}{0.32\textwidth}
        \centering
        \includegraphics[width=1.0\textwidth]{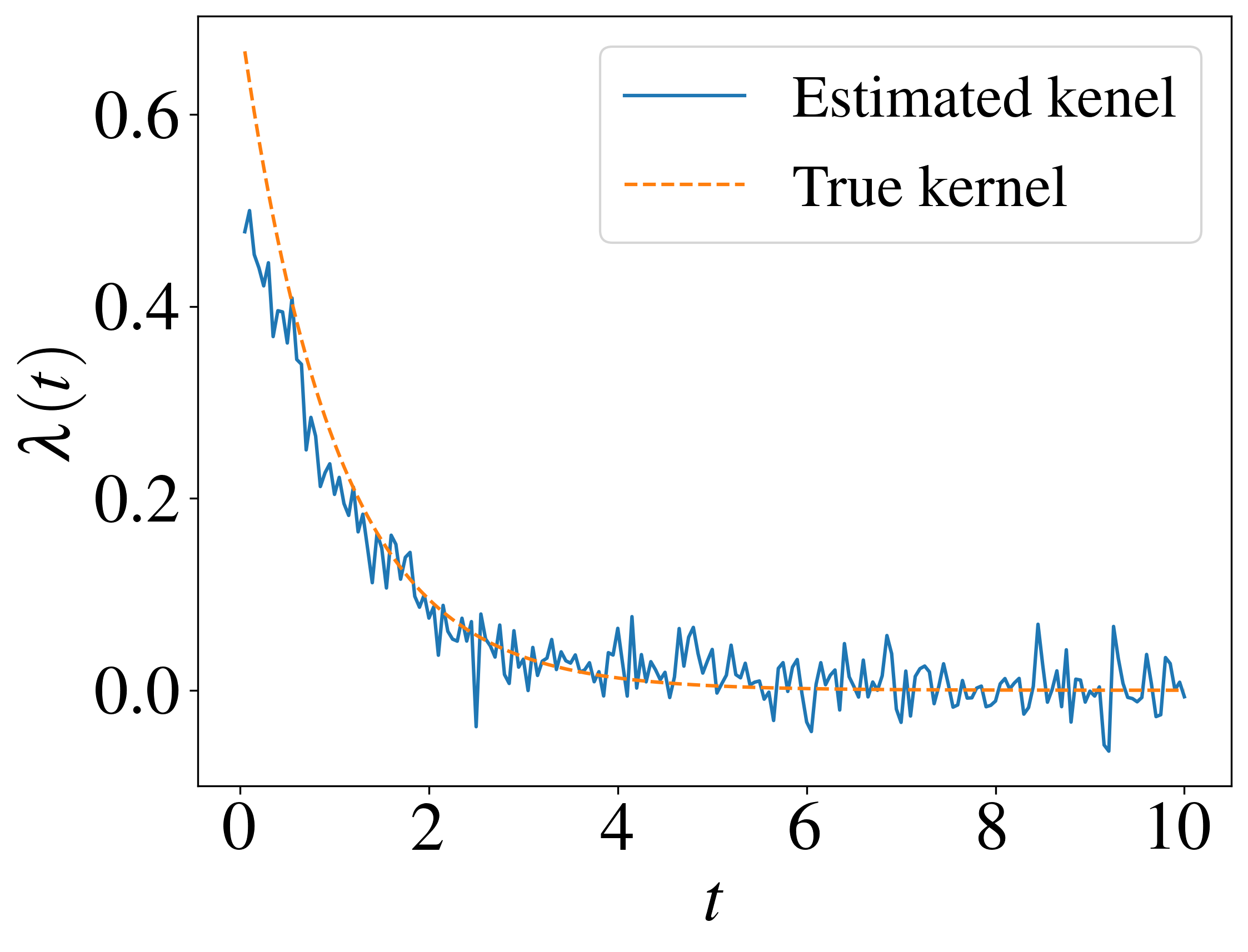}
        \caption{Private ($\sigma^2=10$).}
    \end{subfigure}%
    \caption{Non-private and private estimation of a kernel function $h_{2,2}$ under different noise level. Models are estimated using differentially private conditional gradient algorithm.}
    \label{fig:kernel-fw}
\end{figure*}

\begin{figure*}[t!]
    \centering
    \begin{subfigure}{0.34\textwidth}
        \centering
        \includegraphics[width=1.0\textwidth]{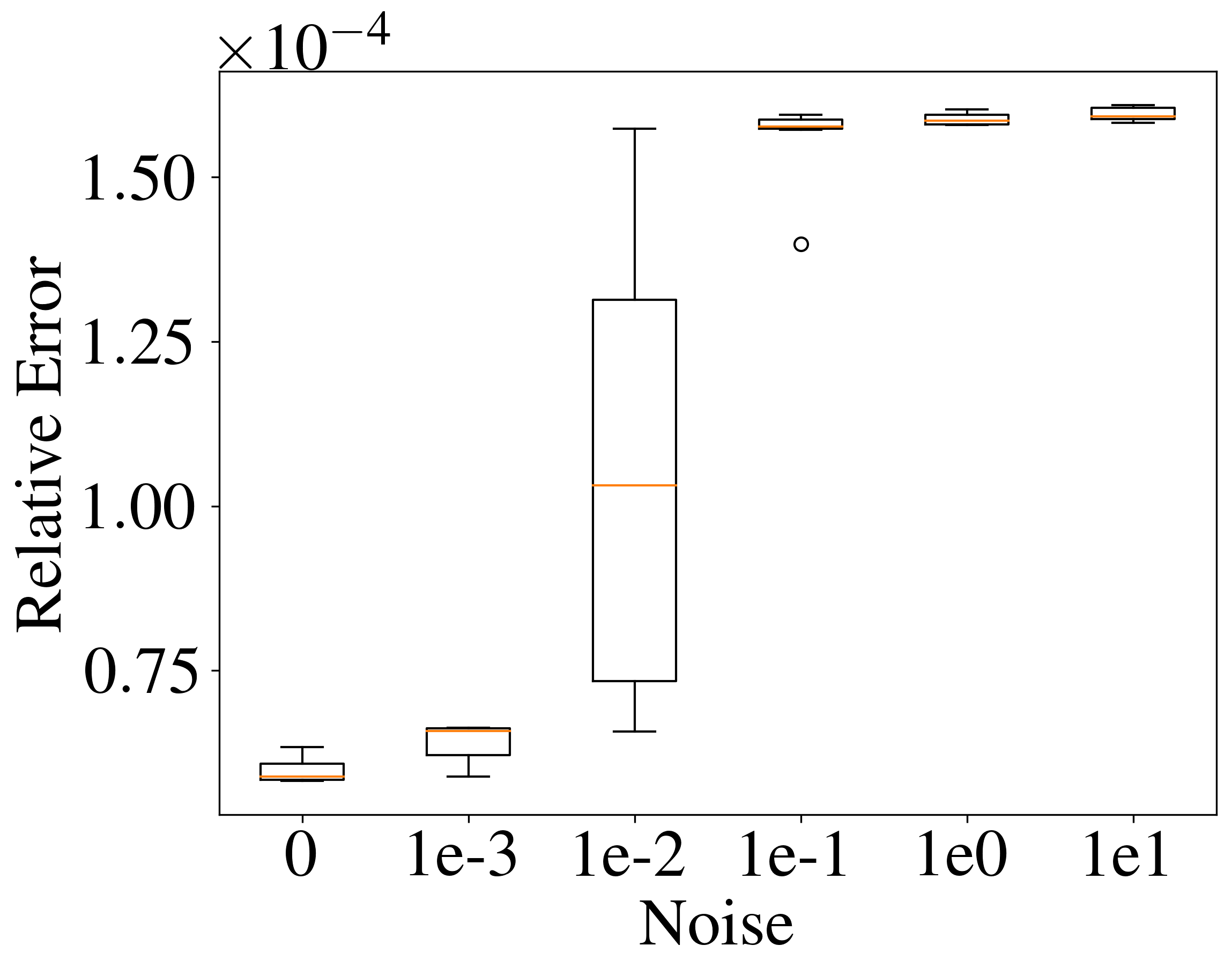}
        \caption{$\Delta=0.01$.}
    \end{subfigure}%
    \begin{subfigure}{0.31\textwidth}
        \centering
        \includegraphics[width=1.0\textwidth]{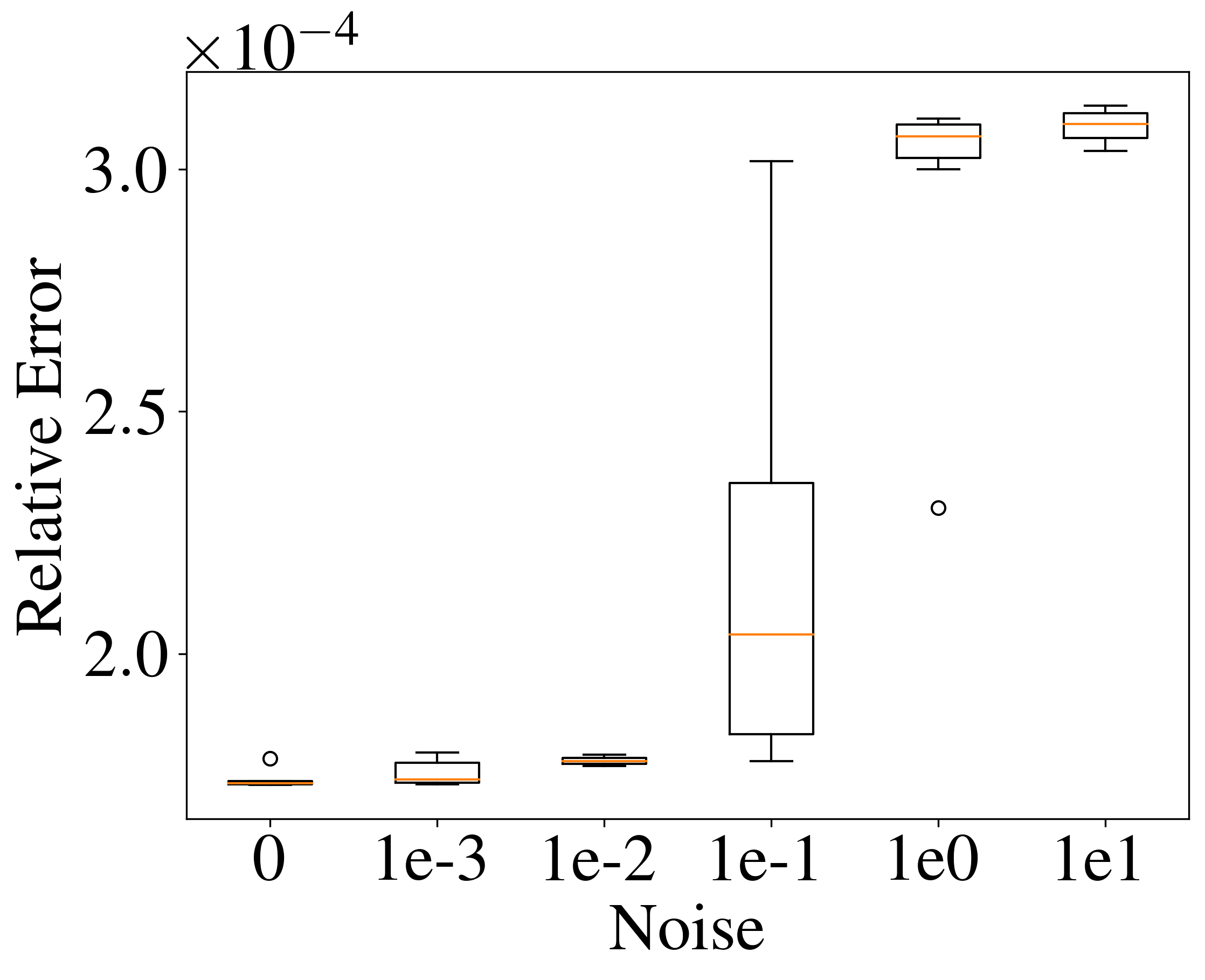}
        \caption{$\Delta=0.03$.}
    \end{subfigure}%
    \begin{subfigure}{0.31\textwidth}
        \centering
        \includegraphics[width=1.0\textwidth]{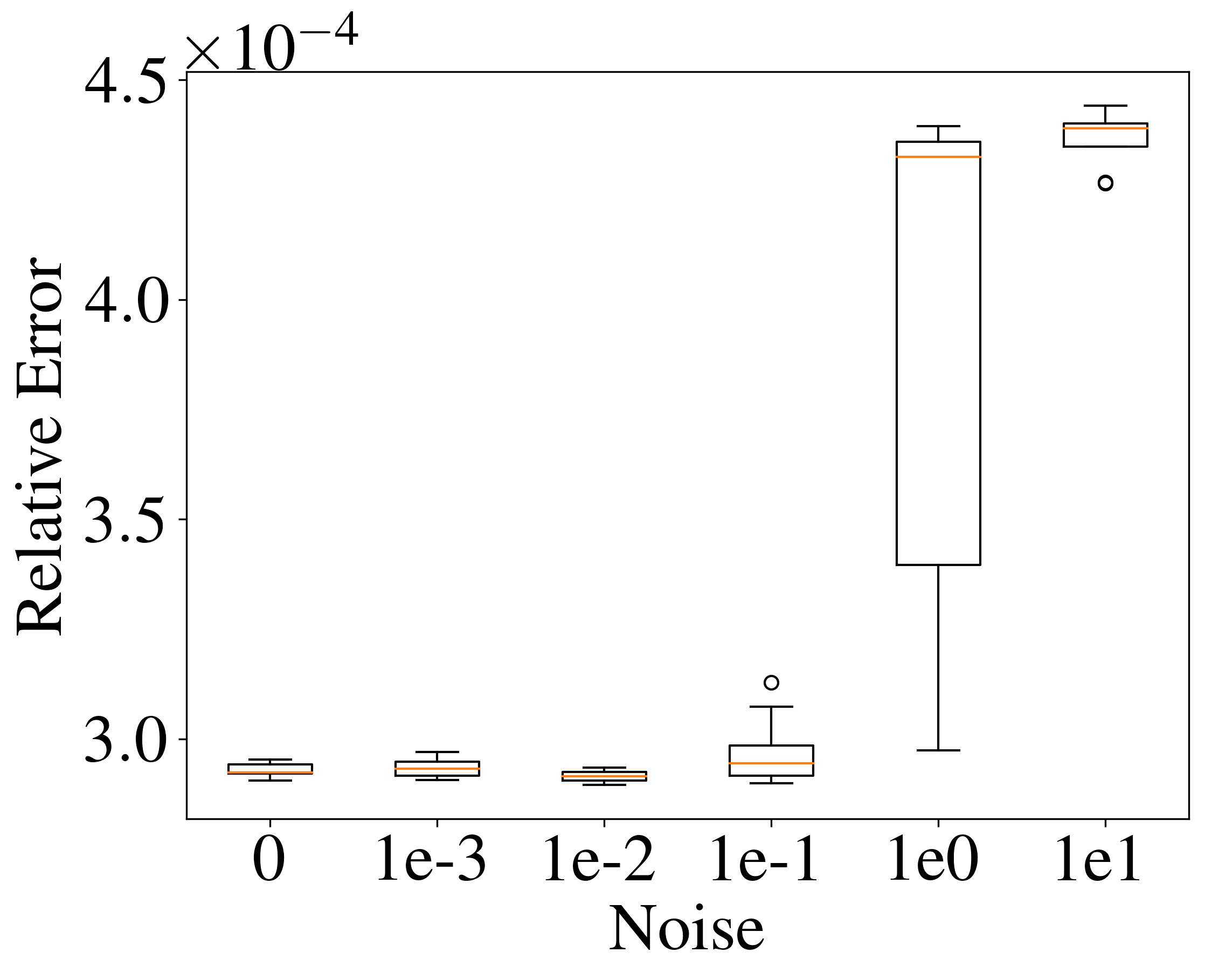}
        \caption{$\Delta=0.05$.}
    \end{subfigure}%
    \caption{Relative error of differentially private estimation under different noise level ($\sigma^2$) using different bin size. Each boxplot is produced from 10 runs. Here \textit{Noise=0} indicates a non-private estimation. Models are estimated using differentially private conditional gradient algorithm.}
    \label{fig:bin-fw}
\end{figure*}

Figure~\ref{fig:kernel-fw} demonstrates the relationship between estimation and noise level. Similar to the case in Figure~\ref{fig:kernel}, our algorithm can still make reasonable estimation under an excessive amount of noise.
Figure~\ref{fig:bin-fw} illustrates the relationship between bin size and relative error. Similar to Figure~\ref{fig:bin}, notice that the estimation quality improves when we apply a smaller bin size, and the differentially private estimation only moderately influences estimation accuracy.
We remark that the differentially private conditional gradient method is robust to noise with small magnitude. For example, in Figure~\ref{fig:bin-fw}, estimation accuracy changes marginally from the noiseless estimation when applying a small noise ($\sigma^2 \leq 10^{-1}$) for $\Delta=0.05$.

\subsection{Experiments on MIMIC-II}

We conduct an additional set of experiments on the MIMIC-II dataset. This dataset contains sensitive information about patients’ visits to hospital. Therefore, privacy protection is needed. We fit a Hawkes process model on noiseless and noisy ($\sigma^2=0.1$) input data.

Table~\ref{tab:mimic} summarizes the RMSE (root mean square error) for next event prediction. From the results, we see that the estimation quality under noisy inputs is on par with that under clean inputs, e.g., the RMSE is $5.96$ and $5.97$ for differentially private estimation using projected gradient descent and conditional gradient, respectively; while the RMSE is $5.92$ for estimation using clean inputs.
These results indicate the practicality of our proposed algorithms in real-world applications.

\begin{table}[ht!]
\vspace{-0.1in}
\caption{Experimental results on the MIMIC-II dataset. Here \textit{PGD} means differentially private projected gradient descent (Algorithm~\ref{alg:DP-GD}), and \textit{CG} means differentially private conditional gradient (Algorithm~\ref{alg:DP-FW}).}
\centering
\begin{tabular}{l|c}
\toprule
Method & RMSE \\ \midrule
Hawkes &  5.92 \\
DP-Hawkes-PGD (Algorithm~\ref{alg:DP-GD}) & 5.96 \\
DP-Hawkes-CG (Algorithm~\ref{alg:DP-FW}) &  5.97 \\
\bottomrule
\end{tabular}
\label{tab:mimic}
\vspace{-0.1in}
\end{table}

%% file: 0-conclusion.tex
\section{Conclusion and Discussion}
\label{sec:conclusion}

We study the problem of estimating Hawkes process models in a differentially private manner.
We discretize the continuous event stream data into bins counts, and approximate the corresponding bin-count sequence using integer-valued auto-regressive (INAR) models. Using such a discretization approach, we introduce a rigorous definition of privacy for event stream data. Then we propose two differentially private algorithms under two different settings, respectively. Theoretically, we establish privacy and utility guarantees for both the algorithms.

One advantage of the proposed discretization approach is that estimation of the INAR model solves a convex least-squares problem, which is in general easier than finding a maximum-likelihood estimator. We remark that there is anther approach~\citep{bacry2015sparse} that also employs a least-squares approach. However, the said approach uses functional least-squares and does not discretize the event stream data, such that privacy cannot be properly defined.



%% file: 0-appendix.tex
\section{Proof of Theorem ~\ref{thm:gd_privacy}} \label{app:proof-privacy-gd}
\begin{proof}
For any $k<K,$ let $Q_k(X) = \nabla \cL(U_k;X) +\zeta_k$ be a random variable defined over the randomness of $\zeta_k$ and conditioned on $U_k.$ Denote $\mu_{X}^k(\cdot)$ be the measure of the random variable $Q_k(X).$ For any neighboring data sets $X$ and $X',$ we define the privacy loss to be $W_t = \big|\log\frac{\mu_{X}(Q_k(X))}{\mu_{X'}(Q_k(X))}\big|$. Since both $\mu_X$ and $\mu_{X'}$ are Gaussian density with covariance matrix $\sigma^2\mathbf{1}_{d\times (dp+1)},$ we have
\begin{align*}
    \frac{\mu_{X}(Q_k(X))}{\mu_{X'}(Q_k(X))}&=\frac{\exp(-\frac{\norm{Q_k(X) -\nabla\cL(U_k;X) }_{\mathrm{F}}^2}{2\sigma^2})}{\exp(-\frac{\norm{Q_k(X) -\nabla\cL(U_k;X') }_{\mathrm{F}}^2}{2\sigma^2})}\\
    &=\exp\left(\frac{\norm{\zeta_k + \nabla\cL(U_k;X)-\nabla\cL(U_k;X')}_{\mathrm{F}}^2-\norm{\zeta_k}_{\mathrm{F}}^2}{2\sigma^2 }\right)\\
    & = \exp\left(\frac{ \norm{\nabla\cL(U_k;X)-\nabla\cL(U_k;X')}_{\mathrm{F}}^2 + 2 <\zeta_k,\nabla\cL(U_k;X)-\nabla\cL(U_k;X')>}{2\sigma^2 }\right)\\
    &\leq  \exp\left(\frac{ 4L_2^2 + 2 <\zeta_k,\nabla\cL(U_k;X)-\nabla\cL(U_k;X')>}{2\sigma^2 }\right).
\end{align*}
The last inequality comes from the fact $\cL$ is $L_2-$Lipschitz. Note that $<\zeta_k,\nabla\cL(U_k;X)-\nabla\cL(U_k;X')>$ follows $N(0,\sigma^2\norm{\nabla\cL(U_k;X)-\nabla\cL(U_k;X')}_{\mathrm{F}}^2).$ The concentration bound of Gaussian distribution leads to the following inequality.
$$\PP[|<\zeta_k,\nabla\cL(U_k;X)-\nabla\cL(U_k;X')>|>2\sigma L_2t]\leq \exp(-\frac{t^2}{2}).$$
Setting $t = \sqrt{2\log \frac{K}{\delta}},$ we then have with probability at least $1-\delta$
\begin{align*}
      \frac{\mu_{X}(Q_k(X))}{\mu_{X'}(Q_k(X))}\leq \exp(\frac{1}{2\sigma^2} (4L_2^2 + 4\sigma L_2 \sqrt{2\log \frac{K}{\delta}})),
\end{align*}
for all $k<K.$ Note that under Assumption \eqref{ass:bounded}, we have $L_2\leq 2 (\Delta B) R$. We then know that with probability at least $1-\delta$ and our choice of $\sigma,$ we have
$$W_t\leq \frac{\epsilon}{2\sqrt{K\log(1/\delta)}}.$$
We can then apply the strong composition theorem \citep{dwork2010boosting}. 
\begin{lemma}[Strong Composition Theorem \citep{dwork2010boosting}] \label{lem:strong_composition}
Let $\epsilon,\delta'\geq 0.$ The class of $\epsilon-$differentially private algorithms satisfies $(\epsilon',\delta')-$differential privacy under $T-$fold adaptive composition for $\epsilon' = \sqrt{2T\ln(1/\delta')}\epsilon + T\epsilon(e^\epsilon -1).$
\end{lemma}
By Lemma \ref{lem:strong_composition}, we can show that with probability at least $1-\delta,$ the privacy loss $W=\sum_{k=0}^{K-1} W_k$ is at most $\epsilon.$ We finish the proof. 
\end{proof}
\section{Proof of Theorem~\ref{thm:gd_utility}}
\label{app:proof-utility-gd}

\begin{proof}
Let $G_k = \nabla \cL(U_k;X) +\cN(0,\sigma^2\mathbf{1}_{d\times (dp+1)})$.
We know that given $U_k$, $\EE[G_k] = \nabla \cL(U_k;X)$. Additionally, we have the following bound on  $\EE[\norm{G_k}_\mathrm{F}^2]$:
\begin{align*}
	\EE[\norm{G_k}_\mathrm{F}^2]
	&= \EE[\norm{\nabla \cL(U_k;X) }_\mathrm{F}^2]
	+ \EE[\norm{\cN(0,\sigma^2\mathbf{1}_{d\times (dp+1)})}_\mathrm{F}^2] \\
	&+ 2\EE[\langle\nabla \cL(U_k;X), \cN(0,\sigma^2\mathbf{1}_{d\times (dp+1)})\rangle]
	\leq  L_2^2 + d(dp+1)\sigma^2.
\end{align*}
We can then apply Theorem 2 in \cite{shamir2013stochastic}. 
\begin{lemma}[Theorem 2 \citep{shamir2013stochastic}]
Suppose the loss $F$ is convex and for some constants $D$ and $G$, it holds that the stochastic gradient $\hat g_t$ satisfies $E[\norm{\hat g_t}]\leq G^2$ for all $t$ and $\sup_{w,w'\in\cW}\norm{w-w'}\leq D.$ Consider SGD with step size $\eta_t = c/\sqrt{t}$ where $c>0$ is a constant. Then for any $T>1,$ it holds that 
$$\EE[F(w_t)-F(w^*)]\leq (\frac{D^2}{c}+cG^2)\frac{2+\log(T)}{\sqrt{T}}.$$
\end{lemma}

That is, we know that if we take $\eta_k = \frac{\Delta B}{\sqrt{k(L_2^2 + d(dp+1)\sigma^2)}},$ we have 
\begin{align*}
    \EE[\cL(U_{K};X)- \min_U\cL(U;X)]
    &= \cO\left(\frac{\Delta B\sqrt{L_2^2 + d(dp+1)\sigma^2}\log K}{\sqrt{K}}\right) \\
	&= \cO\left(\frac{(\Delta B)^2\sqrt{ d(dp+1)}}{\epsilon}\log\frac{K}{\delta}\log K\right).
\end{align*}
Take $K =\cO(1),$ 
we have  
\begin{align*}
	\EE[\cL(U_{K};X)- \min_U\cL(U;X)]
	= \cO\left(\frac{(\Delta B)^2\sqrt{ d(dp+1)}}{\epsilon}\log\frac{1}{\delta}\right).
\end{align*}
This concludes the proof.
\end{proof}

\section{Proof of Theorem~\ref{thm:privacy_FW}}
\begin{proof}
Let $L_2$ be the Lipschitz constant of $\cL$ \eqref{opt:low_rank}. Under Assumption \ref{ass:bounded}, we can show that $L_2\leq 2 (\Delta r) R$.
Applying Theorem B.1 in \citet{talwar2015nearly}, we know that if we take
\begin{align*}
  & \sigma^2= \frac{32L_2^2 K\log^2\frac{K}{\delta}}{\epsilon^2}
  = \frac{128(\Delta r)^2 R^2K\log^2\frac{K}{\delta}}{\epsilon^2},  
\end{align*}
Algorithm \ref{alg:DP-FW} is $(\epsilon,\delta)$ differential private.
\end{proof}
\section{Proof of Theorem~\ref{thm:utility_FW}}
\label{app:proof-utility-fw}




\begin{proof}
The proof of utility follows \citet{jaggi2013revisiting}. By the definition of $\tilde U_k,$ we know that 
\begin{align*}
	&\langle\tilde U_k,\nabla \cL (U_k;X)+\boldsymbol{\zeta}_k(\tilde U_k) \rangle
	\leq \min_{U \in \cC} \langle U, \nabla \cL (U_k;X)+ \boldsymbol{\zeta}_k(U) \rangle.
\end{align*}
Therefore,
\begin{align*}
	&\langle \tilde U_k,\nabla \cL (U_k;X)\rangle
	\leq \min_{U \in \cC}\langle U,\nabla \cL(U_k)+ \boldsymbol{\zeta}_k(U) \rangle
	-\langle\tilde U_k, \boldsymbol{\zeta}_k(\tilde U_k) \rangle \\
	&\hspace{1.075in} \leq  \min_{U \in \cC}\langle U,\nabla \cL (U_k) \rangle + 2\sup_{U \in \cC} |\langle U , \boldsymbol{\zeta}_k(U) \rangle|.
\end{align*}
Note that $U_{k+1} = U_k + \mu(\tilde U_k -U_k).$ By the definition of the curvature constant $\Gamma_{\cL},$ we have
\begin{align}\label{eq:fw-proof}
\cL(U_{k+1})	&= \cL( U_k + \mu(\tilde U_k -U_k))\nonumber\\
&\leq \cL(U_{k}) +\mu\langle\tilde U_k -U_k, \nabla\cL(U_{k})  \rangle +\frac{\mu^2}{2}\Gamma_{\cL}\nonumber\\
&\leq  \cL(U_{k}) +\mu\left[ \min_{U \in \cC}\langle U,\nabla \cL (U_k) \rangle -\langle U_k, \nabla\cL(U_{k})  \rangle\right] + 2\mu\sup_{U \in \cC} |\langle U , \boldsymbol{\zeta}_k(U) \rangle|+\frac{\mu^2}{2}\Gamma_{\cL}\nonumber\\
&\leq  \cL(U_{k}) - \mu g(U_k)+ 2\mu\sup_{U \in \cC} |\langle U , \boldsymbol{\zeta}_k(U) \rangle|+\frac{\mu^2}{2}\Gamma_{\cL},
\end{align}
where $g(U) = \max_{s\in \cC}\langle U -s,\nabla\cL(U)\rangle.$
Note that given the convexity of  $\cL,$ we can verify that $g(U)\geq \cL(U) - \min_{U\in \cC}\cL(U).$ The expectation of $\sup_{U \in \cC} |\langle U , \xi_t(U) \rangle|$ is actually the Gaussian width of $\cC,$ as defined in Definition \ref{def:gaussian_width}. 
	
By the definition of Gaussian width and take expectation on both sides of \eqref{eq:fw-proof}, we have
\begin{align*}
	\EE\cL(U_{k+1})&\leq \EE \cL(U_{k}) - \mu \EE g(U_k)+ 2\mu\sigma\omega(\cC)+\frac{\mu^2}{2}\Gamma_{\cL}\\
	&\leq  \EE \cL(U_{k}) - \mu \left(\EE \cL(U_{k}) - \min_{U\in\cC}\cL(U)\right)+2\mu\sigma\omega(\cC)+\frac{\mu^2}{2}\Gamma_{\cL},
\end{align*}
which directly implies that
\begin{align*}
	 & \EE[ \cL(U_{k+1}) - \min_{U\in\cC}\cL(U)]\leq(1-\mu)  \EE[ \cL(U_{k}) -  \min_{U\in\cC}\cL(U)]+2\mu\sigma\omega(\cC)+\frac{\mu^2}{2}\Gamma_{\cL}.
\end{align*}
By induction, we can then show
\begin{align*}
   \EE[ \cL(U_{K}) - \min_{U\in\cC}\cL(U)]\leq \frac{2\Gamma_{\cL}}{K+2}+8\sigma \omega(\cC). 
\end{align*}

We next utilize  Proposition 10.3 in \cite{vershynin2015estimation}.
\begin{lemma}[Proposition 10.3 \cite{vershynin2015estimation}] 
Consider the unit ball in the space of $d_1\times d_2$ matrices corresponding to the nuclear norm:
$$B_* = \{X\in \RR^{d_1\times d_2}: \norm{X}_*\leq 1\}.$$
Then $$\omega(B_*) \leq  2(\sqrt{d_1}+\sqrt{d_2}).$$
\end{lemma}
Therefore, we know that
\begin{align*}
  & \omega(\cC)\leq 2(\sqrt{dp+1}+ \sqrt{d})\leq 4 \sqrt{dp+1}.  
\end{align*}
Take $\sigma^2 = \cO\left(\frac{KL_2^2\log^2\frac{1}{\delta}}{\epsilon^2}\right) =  \cO\left(\frac{K\Delta^2 r^2 R^2\log^2\frac{K}{\delta}}{\epsilon^2}\right)$, then we have
\begin{align*}
   & \EE[ \cL(U_{K}) - \min_{U\in\cC}\cL(U)]=\cO\left( \frac{\Gamma_{\cL}}{K}+ \frac{\sqrt{K}L_2\sqrt{dp+1}{\log\frac{1}{\delta}}}{\epsilon} \right). 
\end{align*}
Take $K = \left(\frac{\Gamma_{\cL}\epsilon}{L_2\sqrt{dp+1}}\right) ^{\frac{2}{3}},$ we have
\begin{align*}
  & \EE[ \cL(U_{k+1}) - \min_{U\in\cC}\cL(U)]=\cO\left(\frac{\Gamma_{\cL}^{\frac{1}{3}}L_2^{\frac{2}{3}}(dp+1)^{\frac{1}{3}}{\log\frac{1}{\delta}}}{\epsilon^{\frac{2}{3}} }\right) =\cO\left(\frac{R(\Delta r)^{\frac{4}{3}}(dp+1)^{\frac{1}{3}}{\log\frac{1}{\delta}}}{\epsilon^{\frac{2}{3}} }\right).  
\end{align*}
This concludes the proof.
\end{proof}

%% file: main.bbl
\begin{thebibliography}{34}
\expandafter\ifx\csname natexlab\endcsname\relax\def\natexlab#1{#1}\fi
\expandafter\ifx\csname url\endcsname\relax
  \def\url#1{\texttt{#1}}\fi
\expandafter\ifx\csname urlprefix\endcsname\relax\def\urlprefix{}\fi

\bibitem[{Abadi et~al.(2016)Abadi, Chu, Goodfellow, McMahan, Mironov, Talwar
  and Zhang}]{abadi2016deep}
\textsc{Abadi, M.}, \textsc{Chu, A.}, \textsc{Goodfellow, I.}, \textsc{McMahan,
  H.~B.}, \textsc{Mironov, I.}, \textsc{Talwar, K.} and \textsc{Zhang, L.}
  (2016).
\newblock Deep learning with differential privacy.
\newblock In \textit{Proceedings of the 2016 ACM SIGSAC conference on computer
  and communications security}.

\bibitem[{Bacry et~al.(2015{\natexlab{a}})Bacry, Bompaire, Ga{\"\i}ffas and
  Muzy}]{bacry2015sparse}
\textsc{Bacry, E.}, \textsc{Bompaire, M.}, \textsc{Ga{\"\i}ffas, S.} and
  \textsc{Muzy, J.-F.} (2015{\natexlab{a}}).
\newblock Sparse and low-rank multivariate hawkes processes.
\newblock \textit{arXiv preprint arXiv:1501.00725}.

\bibitem[{Bacry et~al.(2017)Bacry, Bompaire, Ga{\"\i}ffas and
  Poulsen}]{bacry2017tick}
\textsc{Bacry, E.}, \textsc{Bompaire, M.}, \textsc{Ga{\"\i}ffas, S.} and
  \textsc{Poulsen, S.} (2017).
\newblock Tick: a python library for statistical learning, with a particular
  emphasis on time-dependent modelling.
\newblock \textit{arXiv preprint arXiv:1707.03003}.

\bibitem[{Bacry et~al.(2015{\natexlab{b}})Bacry, Mastromatteo and
  Muzy}]{bacry2015hawkes}
\textsc{Bacry, E.}, \textsc{Mastromatteo, I.} and \textsc{Muzy, J.-F.}
  (2015{\natexlab{b}}).
\newblock Hawkes processes in finance.
\newblock \textit{Market Microstructure and Liquidity}, \textbf{1} 1550005.

\bibitem[{Bassily et~al.(2014)Bassily, Smith and Thakurta}]{bassily2014private}
\textsc{Bassily, R.}, \textsc{Smith, A.} and \textsc{Thakurta, A.} (2014).
\newblock Private empirical risk minimization, revisited.
\newblock \textit{rem}, \textbf{3} 19.

\bibitem[{Boly et~al.(2020)Boly, Cheysson and Lang}]{Boly2020mixing}
\textsc{Boly, O.}, \textsc{Cheysson, F.} and \textsc{Lang, G.} (2020).
\newblock Mixing conditions for multivariate hawkes processes.
\newblock \textit{Technical Report}.

\bibitem[{Cheysson and Lang(2020)}]{cheysson2020strong}
\textsc{Cheysson, F.} and \textsc{Lang, G.} (2020).
\newblock Strong mixing condition for hawkes processes and application to
  whittle estimation from count data.
\newblock \textit{arXiv preprint arXiv:2003.04314}.

\bibitem[{Du et~al.(2016)Du, Dai, Trivedi, Upadhyay, Gomez{-}Rodriguez and
  Song}]{du2016recurrent}
\textsc{Du, N.}, \textsc{Dai, H.}, \textsc{Trivedi, R.}, \textsc{Upadhyay, U.},
  \textsc{Gomez{-}Rodriguez, M.} and \textsc{Song, L.} (2016).
\newblock Recurrent marked temporal point processes: Embedding event history to
  vector.
\newblock In \textit{Proceedings of the 22nd {ACM} {SIGKDD} International
  Conference on Knowledge Discovery and Data Mining, San Francisco, CA, USA,
  August 13-17, 2016} (B.~Krishnapuram, M.~Shah, A.~J. Smola, C.~C. Aggarwal,
  D.~Shen and R.~Rastogi, eds.). {ACM}.

\bibitem[{Dwork et~al.(2006{\natexlab{a}})Dwork, Kenthapadi, McSherry, Mironov
  and Naor}]{dwork2006our}
\textsc{Dwork, C.}, \textsc{Kenthapadi, K.}, \textsc{McSherry, F.},
  \textsc{Mironov, I.} and \textsc{Naor, M.} (2006{\natexlab{a}}).
\newblock Our data, ourselves: Privacy via distributed noise generation.
\newblock In \textit{Annual International Conference on the Theory and
  Applications of Cryptographic Techniques}. Springer.

\bibitem[{Dwork et~al.(2006{\natexlab{b}})Dwork, McSherry, Nissim and
  Smith}]{dwork2006calibrating}
\textsc{Dwork, C.}, \textsc{McSherry, F.}, \textsc{Nissim, K.} and
  \textsc{Smith, A.} (2006{\natexlab{b}}).
\newblock Calibrating noise to sensitivity in private data analysis.
\newblock In \textit{Theory of cryptography conference}. Springer.

\bibitem[{Dwork et~al.(2014)Dwork, Roth et~al.}]{dwork2014algorithmic}
\textsc{Dwork, C.}, \textsc{Roth, A.} \textsc{et~al.} (2014).
\newblock The algorithmic foundations of differential privacy.
\newblock \textit{Foundations and Trends in Theoretical Computer Science},
  \textbf{9} 211--407.

\bibitem[{Dwork et~al.(2010)Dwork, Rothblum and Vadhan}]{dwork2010boosting}
\textsc{Dwork, C.}, \textsc{Rothblum, G.~N.} and \textsc{Vadhan, S.} (2010).
\newblock Boosting and differential privacy.
\newblock In \textit{2010 IEEE 51st Annual Symposium on Foundations of Computer
  Science}. IEEE.

\bibitem[{Hansen et~al.(2015)Hansen, Reynaud-Bouret, Rivoirard
  et~al.}]{hansen2015lasso}
\textsc{Hansen, N.~R.}, \textsc{Reynaud-Bouret, P.}, \textsc{Rivoirard, V.}
  \textsc{et~al.} (2015).
\newblock Lasso and probabilistic inequalities for multivariate point
  processes.
\newblock \textit{Bernoulli}, \textbf{21} 83--143.

\bibitem[{Hawkes(1971)}]{hawkes1971spectra}
\textsc{Hawkes, A.~G.} (1971).
\newblock Spectra of some self-exciting and mutually exciting point processes.
\newblock \textit{Biometrika}, \textbf{58} 83--90.

\bibitem[{Jaggi(2013)}]{jaggi2013revisiting}
\textsc{Jaggi, M.} (2013).
\newblock Revisiting frank-wolfe: Projection-free sparse convex optimization.
\newblock In \textit{International Conference on Machine Learning}. PMLR.

\bibitem[{Jin-Guan and Yuan(1991)}]{jin1991integer}
\textsc{Jin-Guan, D.} and \textsc{Yuan, L.} (1991).
\newblock The integer-valued autoregressive (inar (p)) model.
\newblock \textit{Journal of time series analysis}, \textbf{12} 129--142.

\bibitem[{Kirchner(2016)}]{kirchner2016hawkes}
\textsc{Kirchner, M.} (2016).
\newblock Hawkes and inar processes.
\newblock \textit{Stochastic Processes and their Applications}, \textbf{126}
  2494--2525.

\bibitem[{Kirchner(2017)}]{kirchner2017estimation}
\textsc{Kirchner, M.} (2017).
\newblock An estimation procedure for the hawkes process.
\newblock \textit{Quantitative Finance}, \textbf{17} 571--595.

\bibitem[{Kirchner and Bercher(2018)}]{kirchner2018nonparametric}
\textsc{Kirchner, M.} and \textsc{Bercher, A.} (2018).
\newblock A nonparametric estimation procedure for the hawkes process:
  comparison with maximum likelihood estimation.
\newblock \textit{Journal of Statistical Computation and Simulation},
  \textbf{88} 1106--1116.

\bibitem[{Mei and Eisner(2017)}]{mei2016neural}
\textsc{Mei, H.} and \textsc{Eisner, J.} (2017).
\newblock The neural hawkes process: {A} neurally self-modulating multivariate
  point process.
\newblock In \textit{Advances in Neural Information Processing Systems 30:
  Annual Conference on Neural Information Processing Systems 2017, December
  4-9, 2017, Long Beach, CA, {USA}} (I.~Guyon, U.~von Luxburg, S.~Bengio, H.~M.
  Wallach, R.~Fergus, S.~V.~N. Vishwanathan and R.~Garnett, eds.).

\bibitem[{Mohler and Brantingham(2018)}]{mohler2018privacy}
\textsc{Mohler, G.} and \textsc{Brantingham, P.~J.} (2018).
\newblock Privacy preserving, crowd sourced crime hawkes processes.
\newblock In \textit{2018 International Workshop on Social Sensing
  (SocialSens)}. IEEE.

\bibitem[{Mohler et~al.(2011)Mohler, Short, Brantingham, Schoenberg and
  Tita}]{mohler2011self}
\textsc{Mohler, G.~O.}, \textsc{Short, M.~B.}, \textsc{Brantingham, P.~J.},
  \textsc{Schoenberg, F.~P.} and \textsc{Tita, G.~E.} (2011).
\newblock Self-exciting point process modeling of crime.
\newblock \textit{Journal of the American Statistical Association},
  \textbf{106} 100--108.

\bibitem[{Ogata(1988)}]{ogata1988statistical}
\textsc{Ogata, Y.} (1988).
\newblock Statistical models for earthquake occurrences and residual analysis
  for point processes.
\newblock \textit{Journal of the American Statistical association}, \textbf{83}
  9--27.

\bibitem[{Shamir and Zhang(2013)}]{shamir2013stochastic}
\textsc{Shamir, O.} and \textsc{Zhang, T.} (2013).
\newblock Stochastic gradient descent for non-smooth optimization: Convergence
  results and optimal averaging schemes.
\newblock In \textit{Proceedings of the 30th International Conference on
  Machine Learning, {ICML} 2013, Atlanta, GA, USA, 16-21 June 2013}, vol.~28 of
  \textit{{JMLR} Workshop and Conference Proceedings}. JMLR.org.

\bibitem[{Sheen et~al.(2020)Sheen, Zhu and Xie}]{sheen2020tensor}
\textsc{Sheen, H.}, \textsc{Zhu, X.} and \textsc{Xie, Y.} (2020).
\newblock Tensor kernel recovery for spatio-temporal hawkes processes.
\newblock \textit{arXiv preprint arXiv:2011.12151}.

\bibitem[{Song et~al.(2013)Song, Chaudhuri and Sarwate}]{song2013stochastic}
\textsc{Song, S.}, \textsc{Chaudhuri, K.} and \textsc{Sarwate, A.~D.} (2013).
\newblock Stochastic gradient descent with differentially private updates.
\newblock In \textit{2013 IEEE Global Conference on Signal and Information
  Processing}. IEEE.

\bibitem[{Talwar et~al.(2015)Talwar, Guha~Thakurta and
  Zhang}]{talwar2015nearly}
\textsc{Talwar, K.}, \textsc{Guha~Thakurta, A.} and \textsc{Zhang, L.} (2015).
\newblock Nearly optimal private lasso.
\newblock \textit{Advances in Neural Information Processing Systems},
  \textbf{28}.

\bibitem[{Vershynin(2015)}]{vershynin2015estimation}
\textsc{Vershynin, R.} (2015).
\newblock Estimation in high dimensions: a geometric perspective.
\newblock In \textit{Sampling theory, a renaissance}. Springer, 3--66.

\bibitem[{Walder et~al.(2020)Walder, Hanks and Slavković}]{walder2020privacy}
\textsc{Walder, A.}, \textsc{Hanks, E.~M.} and \textsc{Slavković, A.} (2020).
\newblock Privacy for spatial point process data.

\bibitem[{Wang et~al.(2018)Wang, Zhang, He and Zha}]{wang2018supervised}
\textsc{Wang, L.}, \textsc{Zhang, W.}, \textsc{He, X.} and \textsc{Zha, H.}
  (2018).
\newblock Supervised reinforcement learning with recurrent neural network for
  dynamic treatment recommendation.
\newblock In \textit{Proceedings of the 24th {ACM} {SIGKDD} International
  Conference on Knowledge Discovery {\&} Data Mining, {KDD} 2018, London, UK,
  August 19-23, 2018} (Y.~Guo and F.~Farooq, eds.). {ACM}.

\bibitem[{Yang et~al.(2011)Yang, Long, Smola, Sadagopan, Zheng and
  Zha}]{yang2011like}
\textsc{Yang, S.}, \textsc{Long, B.}, \textsc{Smola, A.~J.}, \textsc{Sadagopan,
  N.}, \textsc{Zheng, Z.} and \textsc{Zha, H.} (2011).
\newblock Like like alike: joint friendship and interest propagation in social
  networks.
\newblock In \textit{Proceedings of the 20th International Conference on World
  Wide Web, {WWW} 2011, Hyderabad, India, March 28 - April 1, 2011}
  (S.~Srinivasan, K.~Ramamritham, A.~Kumar, M.~P. Ravindra, E.~Bertino and
  R.~Kumar, eds.). {ACM}.

\bibitem[{Yang and Zha(2013)}]{yang2013mixture}
\textsc{Yang, S.} and \textsc{Zha, H.} (2013).
\newblock Mixture of mutually exciting processes for viral diffusion.
\newblock In \textit{Proceedings of the 30th International Conference on
  Machine Learning, {ICML} 2013, Atlanta, GA, USA, 16-21 June 2013}, vol.~28 of
  \textit{{JMLR} Workshop and Conference Proceedings}. JMLR.org.

\bibitem[{Zhou et~al.(2013)Zhou, Zha and Song}]{zhou2013learning}
\textsc{Zhou, K.}, \textsc{Zha, H.} and \textsc{Song, L.} (2013).
\newblock Learning social infectivity in sparse low-rank networks using
  multi-dimensional hawkes processes.
\newblock In \textit{Proceedings of the Sixteenth International Conference on
  Artificial Intelligence and Statistics, {AISTATS} 2013, Scottsdale, AZ, USA,
  April 29 - May 1, 2013}, vol.~31 of \textit{{JMLR} Workshop and Conference
  Proceedings}. JMLR.org.

\bibitem[{Zuo et~al.(2020)Zuo, Jiang, Li, Zhao and Zha}]{zuo2020transformer}
\textsc{Zuo, S.}, \textsc{Jiang, H.}, \textsc{Li, Z.}, \textsc{Zhao, T.} and
  \textsc{Zha, H.} (2020).
\newblock Transformer hawkes process.
\newblock In \textit{International Conference on Machine Learning}. PMLR.

\end{thebibliography}
